\documentclass[10pt, letter, onecolumn]{arxiv}

\usepackage{kantlipsum, lipsum}
\usepackage{dm-colors}
\usepackage{amsmath}
\usepackage{pstricks, pst-node}
\usepackage{verbatim}
\usepackage{multirow}
\usepackage{scalerel}
\usepackage{booktabs}
\usepackage{enumitem}
\usepackage{xspace}
\usepackage{bm}
\usepackage{bbm}
\usepackage{mathtools}
\usepackage{soul}
\usepackage{epsfig}
\usepackage{graphicx}
\usepackage{amssymb}
\usepackage{colortbl}
\usepackage{csquotes}
\usepackage{setspace}
\usepackage{colortbl}
\usepackage{tabularx,ragged2e}
\usepackage{placeins}
\usepackage[symbol]{footmisc}
\usepackage[bibstyle=nature,citestyle=numeric-comp,%
            natbib=true,backend=biber,maxbibnames=99,%
            giveninits=false,sorting=none]{biblatex}
\usepackage{nameref}
\usepackage{varioref}
\usepackage[pagebackref=false,breaklinks=false,%
            colorlinks=true,bookmarks=true,citecolor=ourdarkblue,%
            urlcolor=ourdarkblue,linkcolor=ourdarkblue]{hyperref}
\usepackage[noabbrev,capitalize]{cleveref}
\usepackage{etoc}

\graphicspath{{figures/}}

\title{{\Huge{Towards Generalist Biomedical AI}

}}

\author[$\ast$, $\ddagger$, 1]{Tao Tu} 
\author[$\ast$, $\ddagger$, 2]{Shekoofeh Azizi}
\author[2]{\\Danny Driess}
\author[1]{Mike Schaekermann}
\author[1]{Mohamed Amin}
\author[1]{Pi-Chuan Chang}
\author[1]{Andrew Carroll}
\author[1]{\\Chuck Lau}
\author[2]{Ryutaro Tanno}
\author[2]{Ira Ktena}  
\author[2]{Basil Mustafa}
\author[2]{Aakanksha Chowdhery}
\author[1]{Yun Liu}
\author[2]{\\Simon Kornblith} 
\author[2]{David Fleet}
\author[1]{Philip Mansfield} 
\author[1]{Sushant Prakash}
\author[1]{Renee Wong}
\author[1]{Sunny Virmani}
\author[1]{Christopher Semturs}  
\author[2]{S Sara Mahdavi}
\author[1]{Bradley Green}
\author[1]{Ewa Dominowska}
\author[1]{Blaise Aguera y Arcas}
\author[2]{Joelle Barral}
\author[1]{Dale Webster}
\author[1]{Greg S. Corrado}
\author[1]{Yossi Matias}
\author[1]{Karan Singhal}
\author[2]{Pete Florence}  
\author[$\dagger$, $\ddagger$,1]{\\Alan Karthikesalingam}
\author[$\dagger$, $\ddagger$,1]{Vivek Natarajan}  

\affil[1]{Google Research, }
\affil[2]{Google DeepMind }

\renewcommand{\correspondingauthor}[1]{$\ast$~Equal contributions. %
                                       $\dagger$~Equal leadership. \\%
                                       $\ddagger$~Corresponding authors: \{taotu, shekazizi, alankarthi, natviv\}@google.com }

\begin{document}
\begin{refsection}

\begin{abstract}
Medicine is inherently multimodal, with rich data modalities spanning text, imaging, genomics, and more. Generalist biomedical artificial intelligence (AI) systems that flexibly encode, integrate, and interpret this data at scale can potentially enable impactful applications ranging from scientific discovery to care delivery.
To enable the development of these models, we first curate MultiMedBench, a new multimodal biomedical benchmark. MultiMedBench encompasses 14 diverse tasks such as medical question answering, mammography and dermatology image interpretation, radiology report generation and summarization, and genomic variant calling.
We then introduce Med-PaLM Multimodal (Med-PaLM M), our proof of concept for a generalist biomedical AI system. Med-PaLM M is a large multimodal generative model that flexibly encodes and interprets biomedical data including clinical language, imaging, and genomics with the \textit{same set of model weights}. Med-PaLM M reaches performance competitive with or exceeding the state of the art on all MultiMedBench tasks, often surpassing specialist models by a wide margin. We also report examples of zero-shot generalization to novel medical concepts and tasks, positive transfer learning across tasks, and emergent zero-shot medical reasoning.
To further probe the capabilities and limitations of Med-PaLM M, we conduct a radiologist evaluation of model-generated (and human) chest X-ray reports and observe encouraging performance across model scales. In a side-by-side ranking on 246 retrospective chest X-rays, clinicians express a pairwise preference for Med-PaLM M reports over those produced by radiologists in up to 40.50\% of cases, suggesting potential clinical utility.
While considerable work is needed to validate these models in real-world use cases, our results represent a milestone towards the development of generalist biomedical AI systems.

\end{abstract}

\maketitle


\section{Introduction}
\label{sec:introduction}

Medicine is a multimodal discipline. Clinicians routinely interpret data from a wide range of modalities including clinical notes, laboratory tests, vital signs and observations, medical images, genomics, and more when providing care.

Despite significant progress in biomedical AI, most models today are unimodal single task systems~\cite{esteva2017dermatologist, gulshan2016development, tomavsev2019clinically}. Consider an existing AI system for interpreting mammograms~\cite{mckinney2020international}. Although the system obtains state-of-the-art (SOTA) performance on breast cancer screening, it cannot incorporate relevant information such as patient health records (e.g., breast cancer gene screening status), other modalities such as MRI, or published medical literature that might help contextualize, refine, and improve performance. Further, the system's output is constrained to a pre-specified set of possible classifications. It cannot verbally explain its prediction or engage in a collaborative dialogue to learn from a physician's feedback. This bounds performance and utility of these narrow, single-task, unimodal, specialist AI systems in real-world applications.

\begin{figure*}[t]
\small
    \centering
    \includegraphics[width=0.47\textwidth]{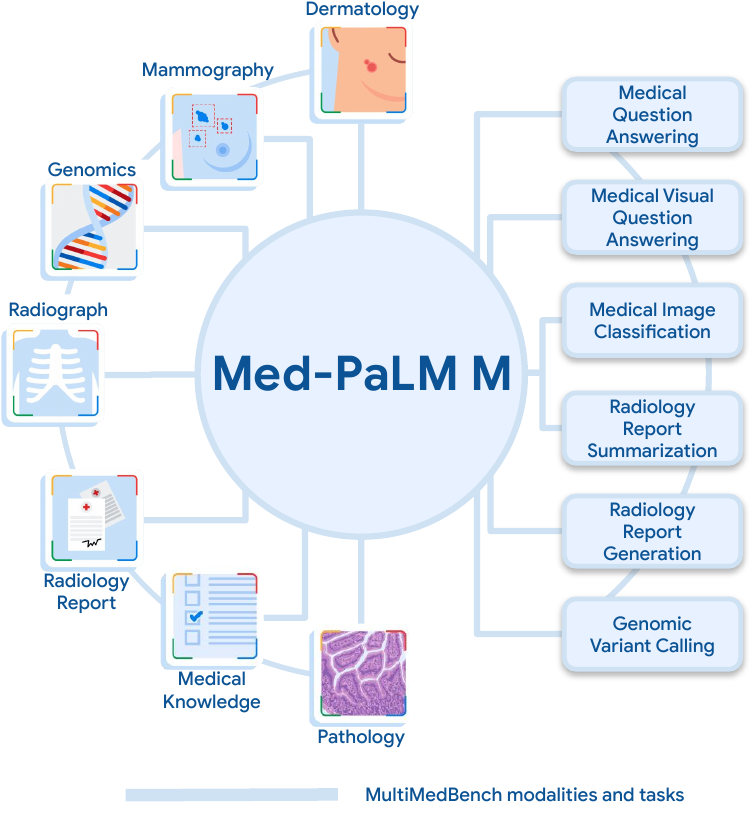} 
    \hfill
    \includegraphics[width=0.45\textwidth]{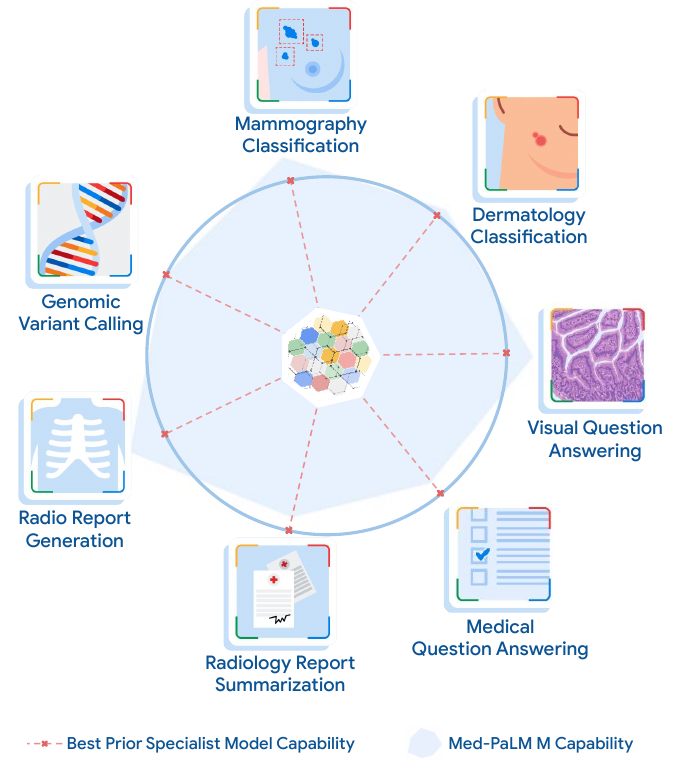} 
    \vspace{6pt} 
    \caption{\textbf{Med-PaLM M overview.} A generalist biomedical AI system should be able to handle a diverse range of biomedical data modalities and tasks. To enable progress towards this overarching goal, we curate MultiMedBench, a benchmark spanning 14 diverse biomedical tasks including question answering, visual question answering, image classification, radiology report generation and summarization, and genomic variant calling. Med-PaLM Multimodal (Med-PaLM M), our proof of concept for such a generalist biomedical AI system (denoted by the shaded blue area) is competitive with or exceeds prior SOTA results from specialists models (denoted by dotted red lines) on all tasks in MultiMedBench. Notably, Med-PaLM M achieves this using a single set of model weights, without any task-specific customization.}
    \vspace{-0pt}
    \label{fig:contributions-overview}
\end{figure*}

The emergence of foundation models~\cite{bommasani2021opportunities} offers an opportunity to rethink the development of medical AI systems. These models are often trained on large-scale data with self-supervised or unsupervised objectives and can be rapidly and effectively
adapted to many downstream tasks and settings using in-context learning or few-shot finetuning~\cite{brown2020language,azizi2023robust}.
Further, they often have impressive generative capabilities that can enable effective human-AI interaction and collaboration. These advances enable the possibility of building a unified biomedical AI system that can interpret multimodal data with complex structures 
to tackle many challenging tasks. As the pace of biomedical data generation and innovation increases, so will the potential impact of such models, with a breadth of possible downstream applications spanning fundamental biomedical discovery to care delivery.

In this work, we detail our progress towards such a \textit{generalist} biomedical AI system - a unified model that can interpret multiple biomedical data modalities and handle many downstream tasks with the \textit{same set of model weights}. 
One of the key challenges of this goal has been the absence of comprehensive multimodal medical benchmarks.
To address this unmet need, we curate MultiMedBench, an open source multimodal medical benchmark spanning language, medical imaging, and genomics modalities with 14 diverse biomedical tasks including question answering, visual question answering, medical image classification, radiology report generation and summarization, and genomic variant calling.

We leverage MultiMedBench to design and develop Med-PaLM Multimodal (Med-PaLM M), a large-scale generalist biomedical AI system building on the recent advances in language~\cite{chowdhery2022palm, singhal2022large} and multimodal foundation models~\cite{driess2023palme, chen2022pali}. In particular, Med-PaLM M is a flexible multimodal sequence-to-sequence architecture that can easily incorporate and interleave various types of multimodal biomedical information. Further, the expressiveness of the modality-agnostic language decoder enables the handling of various biomedical tasks in a simple generative framework with a unified training strategy.

To the best of our knowledge, Med-PaLM M is the first demonstration of a generalist biomedical AI system that can interpret multimodal biomedical data and handle a diverse range of tasks with a single model. Med-PaLM M reaches performance competitive with or exceeding the state-of-the-art (SOTA) on all tasks in MultiMedBench, often surpassing specialized domain and task-specific models by a large margin. In particular, Med-PaLM M exceeds prior state-of-the-art on chest X-ray (CXR) report generation (MIMIC-CXR dataset) by over 8\% on the common success metric (micro-F1) for clinical efficacy. On one of the medical visual question answering tasks (Slake-VQA~\cite{liu2021slake}) in MultiMedBench, Med-PaLM M outperforms the prior SOTA results by over 10\% on the BLEU-1 and F1 metrics.

We perform ablation studies to understand the importance of scale in our generalist multimodal biomedical models and observe significant benefits for tasks that require higher-level language capabilities, such as medical (visual) question answering. Preliminary experiments also suggest evidence of zero-shot generalization to novel medical concepts and tasks across model scales, and emergent capabilities~\cite{wei2022emergent} such as zero-shot multimodal medical reasoning. We further perform radiologist evaluation of AI-generated chest X-ray reports and observe encouraging results across model scales.

Overall, these results demonstrate the potential of generalist biomedical AI systems for medicine. However, significant work remains in terms of large-scale biomedical data access for training such models, validating performance in real world applications, and understanding the safety implications. We outline these key limitations and directions of future research in our study. To summarize, our key contributions are as follows:

\begin{itemize}
    \item \textbf{Curation of MultiMedBench} We introduce MultiMedBench, a new multimodal biomedical benchmark spanning multiple modalities including medical imaging, clinical text and genomics with 14 diverse tasks for training and evaluating generalist biomedical AI systems.
    \item \textbf{Med-PaLM M, the first demonstration of a generalist biomedical AI system} We introduce Med-PaLM M, a single multitask, multimodal biomedical AI system that can perform medical image classification, medical question answering, visual question answering, radiology report generation and summarization, genomic variant calling, and more with the same set of model weights. Med-PaLM M reaches performance competitive with or exceeding state-of-the-art (SOTA) specialist models on multiple tasks in MultiMedBench without any task-specific customization.
    \item \textbf{Evidence of novel emergent capabilities in Med-PaLM M} Beyond quantitative evaluations of task performance, we observe evidence of zero-shot medical reasoning, generalization to novel medical concepts and tasks, and positive transfer across tasks. These experiments suggest promising potential of such systems in downstream data-scarce biomedical applications.
   \item \textbf{Human evaluation of Med-PaLM M outputs} Beyond automated metrics, we perform radiologist evaluation of chest X-ray reports generated by Med-PaLM M across different model scales.
   In a blinded side-by-side ranking on 246 retrospective chest X-rays, clinicians expressed a pairwise preference for Med-PaLM M reports over those produced by radiologists in up to 40.50\% of cases. Furthermore, the best Med-PaLM M model has on average 0.25 clinically significant errors per report.  These results are on par with human baselines from prior work~\cite{jeong2023multimodal}, suggesting potential clinical utility.
\end{itemize}

\section{Related Work}
\label{sec:related-work}

\subsection{Foundation models, multimodality, and generalists}
The emergence of the \textbf{foundation model} paradigm~\cite{bommasani2021opportunities} has had widespread impact across a variety of applications in language~\cite{chowdhery2022palm}, vision~\cite{dehghani2023scaling}, and other modalities~\cite{borsos2023audiolm}.  While the idea of transfer learning ~\cite{caruana1997multitask,thrun1998lifelong} using the weights of pretrained models has existed for decades~\cite{hinton2006fast,bengio2006greedy,vincent2008extracting,bengio2012deep}, a shift has come about due to the scale of data and compute used for pretraining such models~\cite{kaplan2020scaling}.  The notion of a foundation model further indicates that the model can be adapted to a wide range of downstream tasks~\cite{bommasani2021opportunities}.

Within the foundation model paradigm, \textbf{multimodality} \cite{ngiam2011multimodal} has also had a variety of important impacts -- in the datasets~\cite{schuhmann2022laion}, in the inter-modality supervision~\cite{jaegle2021perceiver}, and in the generality and unification of task specification~\cite{tsimpoukelli2021multimodal, alayrac2022flamingo}.  For example, language has specifically been an important enabler of foundation models in other modalities~\cite{chen2022pali, agostinelli2023musiclm}.  Visual foundation models such as CLIP~\cite{radford2021learning} are made possible by training on language-labeled visual datasets~\cite{thomee2016yfcc100m,schuhmann2022laion}, which are easier to collect from large-scale internet data than classification datasets with pre-determined class labels (i.e., ImageNet~\cite{deng2009imagenet}).  The benefits of joint language-and-vision supervision has also been noteworthy in generative modeling of images~\cite{rombach2021highresolution}, where text-to-image generative modeling has been notably more successful at producing high-fidelity image generation~\cite{saharia2022photorealistic} than purely unconditioned generative image modeling~\cite{dhariwal2021diffusion}.  Further, the flexibility of language also enables a wide range of task specifications all via one unified output space~\cite{radford2019language} -- it is possible to phrase tasks traditionally addressed by different output spaces, such as object detection and object classification, all jointly via the output space of language~\cite{chen2021pix2seq}.  Med-PaLM M additionally benefits from the generality of multimodality, both via a model~\cite{driess2023palme} pretrained on large vision-language datasets~\cite{chen2022pali}, and also by further biomedical domain finetuning through a unified generative language output space.

A related notion to that of a foundation model is that of a \textbf{generalist model} -- the same model with the same set of weights, without finetuning, can excel at a wide variety of tasks.  A single multitask \cite{caruana1997multitask} model which can address many tasks has been of long standing interest~\cite{collobert2008unified, ruder2017overview}, including for example in the reinforcement learning community \cite{lee2022multi}. Language-only models such as GPT-3~\cite{brown2020language} and PaLM~\cite{chowdhery2022palm} simultaneously excel at many tasks using only prompting and in-context learning. Recent work has also explored generalist models capable not only of performing many tasks, but also of processing many modalities~\cite{lu2022unified}. For example, the capabilities of Gato~\cite{reed2022generalist} span language, vision, and agent policy learning. PaLM-E~\cite{driess2023palme} further shows that it is possible to obtain a single generalist model which excels at  language-only tasks, vision-language tasks, and embodied vision-language tasks.
Med-PaLM M is specifically a generalist model designed for the biomedical domain, built by finetuning and aligning the PaLM-E generalist model.

\subsection{Multimodal foundation models in biomedicine}

Given the potential, there has been significant interest in multimodal foundation models for different biomedical applications.~\citet{moor2023foundation} discuss the notion of generalist medical AI, albeit without implementation or empirical results.~\citet{theodoris2023transfer} introduce Geneformer, a transformer~\cite{vaswani2017attention} based model pretrained on a corpus of about 30 million single-cell transcriptomes to enable context-specific predictions in low data network biology applications. BiomedGPT~\cite{zhang2023biomedgpt} is a multi-task biomedical foundation model pretrained on a diverse source of medical images, medical literature, and clinical notes using a combination of language model (LM) and masked image infilling objectives.  However, all these efforts are pretrained models and as such they require further task-specific data and finetuning to enable downstream applications. In contrast, Med-PaLM M is directly trained to jointly solve many biomedical tasks at the same time without requiring any further finetuning or model parameter updates. LLaVA-Med~\cite{li2023llava} is perhaps most similar to our effort. The authors use PubMed and GPT-4~\cite{bubeck2023sparks} to curate a multimodal instruction following dataset and finetune a LLaVA model with it. However, the experiments are limited to three medical visual question answering datasets and qualitative examples of conversations conditioned on a medical image. In contrast, our work is more comprehensive, spanning multiple modalities including medical imaging, clinical text, and genomics with 14 diverse tasks and expert evaluation of model outputs.

\subsection{Multimodal medical AI benchmarks}
To the best of our knowledge, there have been limited attempts to curate benchmarks for training and evaluating generalist biomedical AI models. Perhaps the work closest in spirit is BenchMD~\cite{wantlin2023benchmd}. The benchmark spans 19 publicly available datasets and 7 medical modalities, including 1D sensor data, 2D images, and 3D volumetric scans. However, their tasks are primarily focused on classification whereas our benchmark also includes generative tasks such as medical (visual) question answering, radiology report generation and summarization. Furthermore, there is currently no implementation of a generalist biomedical AI system that can competently handle all these tasks simultaneously.

\section{MultiMedBench: A Benchmark for Generalist Biomedical AI}
\label{sec:MultiMedBench}
We next describe MultiMedBench, a benchmark we curated to enable the development and evaluation of generalist biomedical AI. 
MultiMedBench is a multi-task, multimodal benchmark comprising 12 de-identified open source datasets and 14 individual tasks. It measures the capability of a general-purpose biomedical AI to perform a variety of clinically-relevant tasks. The benchmark covers a wide range of data sources including medical questions, radiology reports, pathology, dermatology, chest X-ray, mammography, and genomics. Tasks in MultiMedBench vary across the following axes:

\begin{itemize}
    \item \textbf{Task type:} question answering, report generation and summarization, visual question answering, medical image classification, and genomic variant calling.
    \item \textbf{Modality:} text, radiology (CT, MRI, and X-ray), pathology, dermatology, mammography, and genomics.
    \item \textbf{Output format:} open-ended generation for all tasks including classification.
\end{itemize}

\begin{table}[ht]
\small
\centering
\caption{\textbf{MultiMedBench overview.} Summary of MultiMedBench, the benchmark we introduce for the development and evaluation of Med-PaLM M. MultiMedBench consists of 14 individual tasks across 5 task types and 12 datasets spanning 7 biomedical data modalities. In total, the benchmark contains over 1 million samples.}
\label{tab:multimedbench-overview}
\begin{tabular}{@{}c@{\hspace{.01cm}}@{\hspace{.01cm}}c@{\hspace{.01cm}}ccc}
\toprule
\textbf{Task Type} & \textbf{Modality} & \textbf{Dataset}          & \textbf{Description}  \\ \midrule
\multirow{3}{*}{Question Answering} & \multirow{3}{*}{Text}
& MedQA       & {\begin{tabular}[c]{@{}c@{}}US medical licensing exam-style, multiple-choice\end{tabular}} 
 \\ 
& & MedMCQA     &{\begin{tabular}[c]{@{}c@{}}Indian medical entrance exams, multiple-choice\end{tabular}} \\ 

& & PubMedQA     &{\begin{tabular}[c]{@{}c@{}}Biomedical literature questions, multiple-choice\end{tabular}} \\ 
\midrule
Report Summarization & Radiology & MIMIC-III       &{\begin{tabular}[c]{@{}c@{}}Summarizing findings in radiology reports\end{tabular}}  \\ 
\midrule
\multirow{3}{*}{\begin{tabular}[c]{@{}c@{}}Visual\\Question Answering\end{tabular}} & \multirow{2}{*}{Radiology}
& VQA-RAD & {\begin{tabular}[c]{@{}c@{}}Close/open-ended VQA on radiology images\end{tabular}}  \\
&& Slake-VQA & {\begin{tabular}[c]{@{}c@{}}English-Chinese bilingual VQA on radiology images\end{tabular}} \\
& Pathology & Path-VQA  & {\begin{tabular}[c]{@{}c@{}}Close/open-ended VQA on pathology images\end{tabular}}  \\
\midrule
Report Generation &Chest X-ray &MIMIC-CXR         & {\begin{tabular}[c]{@{}c@{}} Chest X-ray report generation\end{tabular}}  \\ 
\midrule
\multirow{7}{*}{\begin{tabular}[c]{@{}c@{}}Medical\\Image Classification\end{tabular}}
& Chest X-ray & MIMIC-CXR        &  {\begin{tabular}[c]{@{}c@{}} Binary classification of chest X-ray abnormalities\end{tabular}}  \\ 
& Dermatology &PAD-UFES-20  & {\begin{tabular}[c]{@{}c@{}} 6-class skin lesion image classification\end{tabular}} \\
& \multirow{3}{*}{Mammography}
& VinDr-Mammo         & {\begin{tabular}[c]{@{}c@{}} 5-class breast-level BI-RADS classification\end{tabular}} \\
&& CBIS-DDSM         &  {\begin{tabular}[c]{@{}c@{}} 3-class lesion-level classification (mass)\end{tabular}}  
\\
&& CBIS-DDSM          &  {\begin{tabular}[c]{@{}c@{}} 3-class lesion-level classification (calcification)\end{tabular}}
\\
& {\begin{tabular}[c]{@{}c@{}}Genomics\end{tabular}} & {\begin{tabular}[c]{@{}c@{}}PrecisionFDA\\Truth Challenge V2\end{tabular}}     & {\begin{tabular}[c]{@{}c@{}} Genomic variant calling as 3-class image classification\end{tabular}}  \\

\bottomrule 
\end{tabular}
\end{table}

Language-only tasks consist of medical question answering, including three of the MultiMedQA tasks used in \citet{singhal2022large}, and radiology report summarization. They were selected to assess a model's ability to comprehend, recall, and manipulate medical knowledge. Multimodal tasks include medical visual question answering (VQA), medical image classification, chest X-ray report generation, and genomic variant calling, which are well-suited to evaluate both the visual understanding and multimodal reasoning capabilities of these models.~\cref{tab:multimedbench-overview} includes an overview of the datasets and tasks in MultiMedBench - in total, the benchmark contains over 1 million samples. For detailed descriptions of individual datasets and tasks, see~\cref{appendix:multimedbench-details}.

\section{Med-PaLM M: A Proof of Concept for Generalist Biomedical AI}
\label{sec:med-palm-m}
In this section, we detail the methods underpinning the development of the Med-PaLM M model.
We first review preliminaries of the pretrained models in \cref{subsec:model-prelims} from which Med-PaLM M inherits, then discuss the datasets and training details involved in the finetuning and specialization of the model to the biomedical domain \cref{subsec:med-palm-m}.

\subsection{Model preliminaries}\label{subsec:model-prelims}

Note that Med-PaLM M inherits not only the architectures of these pretrained models, but also the general domain knowledge encoded in their model parameters.

\paragraph{Pathways Language Model (PaLM)} introduced by~\citet{chowdhery2022palm} is a densely-connected decoder-only Transformer \cite{vaswani2017attention} based large language model (LLM) trained using Pathways~\cite{barham2022pathways}, a large-scale ML accelerator orchestration system that enables highly efficient training across TPU pods. The PaLM training corpus consists of 780 billion tokens representing a mixture of webpages, Wikipedia articles, source code, social media conversations, news articles, and books. PaLM models were trained at sizes of 8, 62, and 540 billion parameters, and all three PaLM model variants are trained for one epoch of the training data. At the time of its announcement, PaLM 540B achieved breakthrough performance, outperforming finetuned state-of-the-art models on a suite of multi-step reasoning tasks and exceeding average human performance on BIG-bench~\cite{srivastava2022beyond}. 

\paragraph{Vision Transformer (ViT)} introduced by~\citet{dosovitskiy2020image} extends the Transformer~\cite{vaswani2017attention} architecture to visual data such as images and videos. In this work, we consider two ViT pre-trained models as vision encoders, the 4 billion (4B) parameters model from~\citet{chen2022pali} and the 22 billion (22B) parameters model from~\citet{dehghani2023scaling}. Both of these models were pretrained via supervised learning on a large classification dataset \cite{sun2017revisiting,zhai2022scaling} of approximately 4 billion images. 

\paragraph{PaLM-E} introduced by~\citet{driess2023palme} is a multimodal language model that can process sequences of multimodal inputs including text, vision, and sensor signals. The primary PaLM-E model uses pretrained PaLM and ViT, and was initially developed for embodied robotics applications but demonstrated strong performance on multiple vision language benchmarks such as OK-VQA~\cite{marino2019ok} and VQA v2~\cite{goyal2017making}. Furthermore, PaLM-E offers the flexibility to interleave images, text and sensor signals in a single prompt, enabling the model to make predictions with a fully multimodal context. PaLM-E also exhibits a wide array of capabilities including zero-shot multimodal chain-of-thought (CoT) reasoning, and few-shot in-context learning.
We therefore leverage the PaLM-E model as the base architecture for Med-PaLM M.

We consider three different combinations of LLM and vision encoders in our study - PaLM 8B with ViT 4B (PaLM-E 12B), PaLM 62B with ViT 22B (PaLM-E 84B) and PaLM 540B with ViT 22B (PaLM-E 562B). All models were pretrained on diverse vision-language datasets in addition to tasks across multiple robot embodiments as described in~\citet{driess2023palme}.  

\subsection{Putting it all together: Med-PaLM~M}
\label{subsec:med-palm-m}

Med-PaLM M is developed by finetuning and aligning the PaLM-E model to the biomedical domain using MultiMedBench. The following summarizes important methodological details underlying the development of the model.

\paragraph{Dataset and preprocessing}
We resized all the images in MultiMedBench to $224 \times 224\times3$, while preserving the original aspect ratio with padding if needed. The gray-scale images were converted to 3-channel images by stacking up the same image along the channel dimension. Task-specific prepossessing methods such as class balancing and image data augmentation are described in detail for each task in~\cref{appendix:multimedbench-details}.

\begin{figure*}[t]
\small
    \centering
    \includegraphics[width=0.98\textwidth]{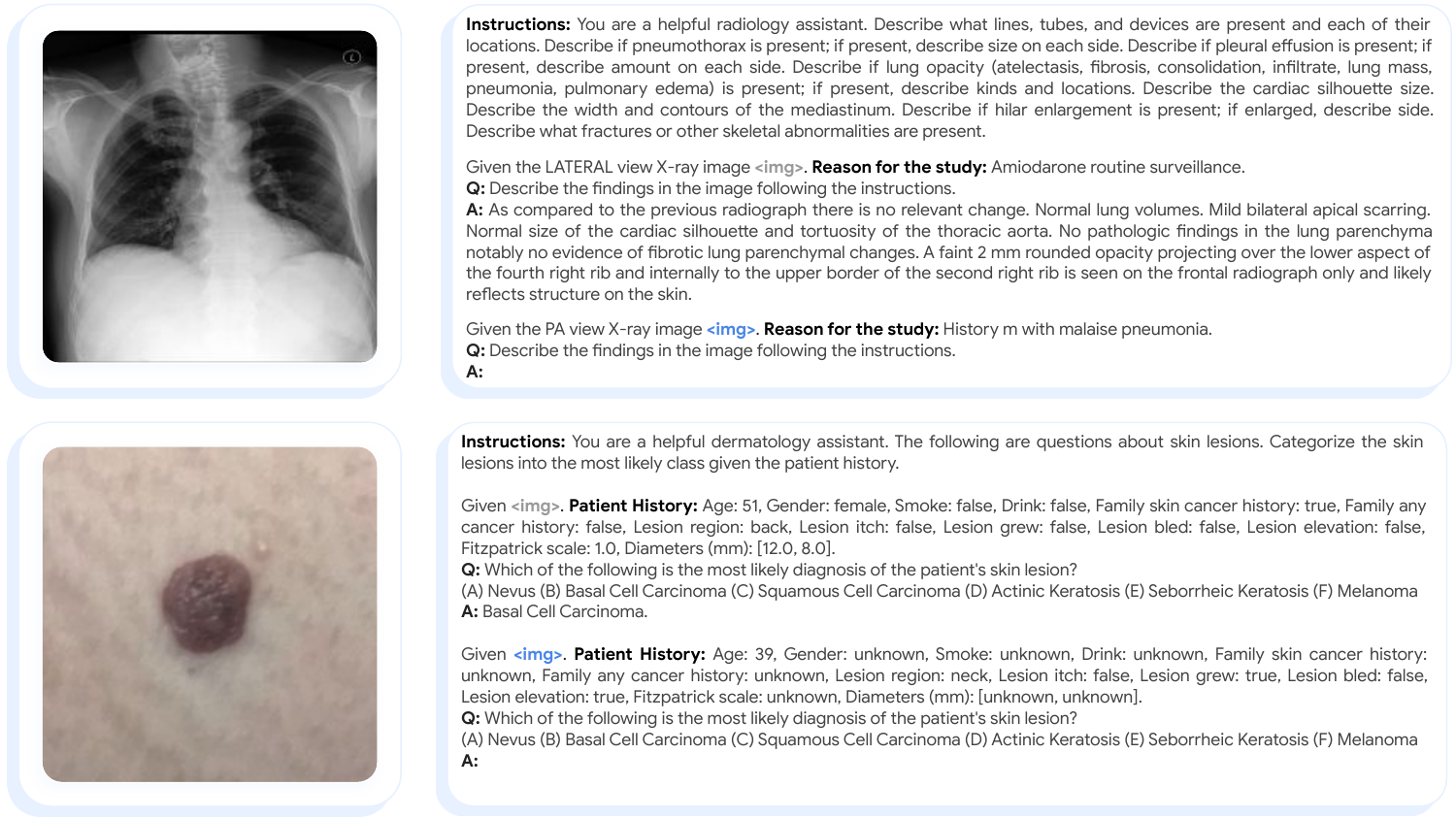} 
    \vspace{5pt}    
    \caption{\textbf{Illustration of instruction task prompting with one-shot exemplar.} (top) shows the task prompt for the chest X-ray report generation task. It consists of task-specific instructions, a text-only ``one-shot exemplar'' (omitting the corresponding image but preserving the target answer), and the actual question. The X-ray image is embedded and interleaved with textual context including view orientation and reason for the study in addition to the question. (bottom) shows the task prompt for the dermatology classification task. We formulate the skin lesion classification task as a multiple choice question answering task with all the class labels provided as individual answer options. Similar to the chest X-ray report generation task, skin lesion image tokens are interleaved with the patient clinical history as additional context to the question. The blue <img> denotes the position in the prompt where the image tokens are embedded.}
    \vspace{-5pt}
    \label{fig:task_prompts}
\end{figure*}

\paragraph{Instruction task prompting and one-shot exemplar} Our goal is to train a generalist biomedical AI model to perform multiple tasks with multimodal inputs using a unified model architecture and a single set of model parameters. To this end, we trained the model with a mixture of distinct tasks simultaneously via instruction tuning \cite{wei2021finetuned}. Specifically, we provided the model with task-specific instructions to prompt the model to perform different types of tasks in a unified generative framework. The task prompt consists of an instruction, relevant context information, and a question. For example, as shown in~\cref{fig:task_prompts}, in the chest X-ray report generation task, we included the reason for the study and the image orientation information as additional context information for the model to condition its prediction on. Similarly, for the dermatology classification task, we provided the patient clinical history associated with the skin lesion image. We formulated all classification tasks as multiple choice questions where all possible class labels are provided as individual answer options and the model was prompted to generate the most likely answer as the target output. For other generative tasks such as visual question answering and report generation and summarization, the model was finetuned on the target response.

In order to enable the model to better follow instructions, for the majority of tasks (see~\cref{tab-appendix:data mixture}), we added a text-only \textit{``one-shot exemplar''} to the task prompt to condition the language model's prediction. The one-shot exemplar helps prompt the model with a partial input-output pair.
Importantly, for multimodal tasks, we replaced the actual image in the exemplar with a dummy text placeholder (with the text string ``<img>''): this (i) preserves training compute efficiency for single-image training, and also (ii) bypasses potential interference from cross-attention between a given text token and image tokens from multiple images~\cite{alayrac2022flamingo}. Our results show that this scheme is effective in prompting the model to generate the desired format of responses as detailed in~\cref{sec:results}.

\paragraph{Model training} We finetuned the pretrained 12B, 84B, and 562B parameter variants of PaLM-E on MultiMedBench tasks with mixture ratios denoted in~\cref{tab-appendix:data mixture}. These mixture ratios were empirically determined such that they are approximately proportional to the number of training samples in each dataset and ensuring at least one sample from each task is present in one batch. We performed an end-to-end finetuning of the PaLM-E model with the entire set of model parameters updated during training. For multimodal tasks, image tokens were interleaved with text tokens to form multimodal context input to the PaLM-E model. 
The multimodal context input contains at most 1 image for all finetuning tasks. However, we note that Med-PaLM M is able to process inputs with multiple images during inference.

We used the Adafactor optimizer~\cite{shazeer2018adafactor} with momentum of $\beta_1=0.9$, dropout rate of 0.1, and a constant learning rate schedule.
We used different sets of hyperparameters in our finetuning experiments for different model sizes, which are further detailed in~\cref{tab-appendix:finetuning hyperparameters}. 

The resulting model, Med-PaLM M (12B, 84B, and 562B), is adapted to the biomedical domain with the capability to encode and interpret multimodal inputs and perform tasks including medical (visual) question answering, radiology report generation and summarization, medical image classification, and genomic variant calling.

\section{Evaluation}
\label{sec:evaluation}

In this section, we describe the purpose, scope, and methods of experimental evaluations. Results are presented in Section~\ref{sec:results}.
Evaluation experiments of Med-PaLM M were designed for the following purposes: 
\begin{itemize}
    \item \textbf{Evaluate generalist capabilities} We evaluated Med-PaLM M on all tasks in MultiMedBench across model scales. We provide initial insights on the effect of scaling ViT and LLM components across different tasks. We compared performance to previous SOTA (including specialist single-task or single-modality methods) and a state-of-art generalist model (PaLM-E) without biomedical finetuning.
    \item \textbf{Explore novel emergent capabilities} One hypothesized benefit of training a single flexible multimodal generalist AI system across diverse tasks is the emergence of novel capabilities arising from language enabled combinatorial generalization, such as to novel medical concepts and tasks. We explored this via qualitative and qualitative experiments.
    \item \textbf{Measure radiology report generation quality} Automatic natural language generation (NLG) metrics do not provide sufficient evaluation of the clinical applicability of AI-generated radiology reports. We therefore performed expert radiologist evaluation of AI-generated reports on the MIMIC-CXR dataset, including comparison to the radiologist-provided reference reports.
\end{itemize}

\subsection{Evaluation on MultiMedBench}
Med-PaLM M was simultaneously finetuned on a mixture of language-only and multimodal biomedical tasks in MultiMedBench. We assessed the model's in-distribution performance on these tasks by comparing to the corresponding SOTA results obtained from separate specialist models. Specifically, we used the same few-shot setup as in training for each task during evaluation. Task-specific metrics were computed on the test split of each task and compared to prior SOTA specialist AI systems. Note that for a small number of tasks described in~\cref{tab:multimedbench-overview}, we were not able to find a sufficiently similar prior attempt for comparison.

\subsection{Evaluation of language enabled zero-shot generalization}

To probe Med-PaLM M's ability to generalize to previously unseen medical concepts, we evaluate the model's ability to predict the presence or absence of tuberculosis (TB) from chest X-ray images.
We used the Montgomery County chest X-ray set (MC) for this purpose.
The dataset contains 138 frontal chest X-rays, of which 80 are normal cases and 58 cases have manifestations of TB~\cite{jaeger2014two}. Each case also contains annotations on the abnormality seen in the lung. We note that Med-PaLM M has been trained on MIMIC-CXR dataset; however, it is not trained to explicitly predict the TB disease label.

We evaluated the accuracy across model scales by formulating this problem as a two-choice question answering task where the model was prompted (with a text-only one-shot exemplar) to generate a yes/no answer about the presence of TB in the input image.

We further explored zero-shot chain-of-thought (CoT) multimodal medical reasoning ability of the model by prompting with a text-only exemplar (without the corresponding image) and prompting the model to generate the class prediction and an accompanying report describing the image findings. We note that while we did prompt the model with a single text-only input-output pair, we omitted the image (used a dummy text placeholder instead) and the text exemplar was hand-crafted rather than drawn from the training set. Hence, this approach can be considered zero-shot rather than one-shot.

In order to assess Med-PaLM M's ability to generalize to novel task scenarios, we evaluated the model performance on two-view chest X-ray report generation - this is a novel task given the model was trained to generate reports only from a single-view chest X-ray.

Finally, we also probed for evidence of positive task transfer as a result of jointly training a single generalist model to solve many different biomedical tasks. To this end, we performed an ablation study where we trained a Med-PaLM M 84B variant by excluding the MIMIC-CXR classification tasks from the task mixture. We compared this model variant to the Med-PaLM M 84B variant trained on the complete MultiMedBench mixture on the chest X-ray report generation task with the expectation of improved performance in the latter. 

\subsection{Clinician evaluation of radiology report generation}
To further assess the quality and clinical applicability of chest X-ray reports generated by Med-PaLM M and understand the effect of model scaling, we conducted a human evaluation using the MIMIC-CXR dataset. The evaluation was performed by four qualified thoracic radiologists based in India.

\paragraph{Dataset} The evaluation set consisted of 246 cases selected from the MIMIC-CXR test split. To match the expected input format of Med-PaLM M, we selected a single image from each study. We excluded studies that had ground truth reports mentioning multiple X-ray views or past examinations of the same patient.

\paragraph{Procedure} We conducted two complementary human evaluations: (1) \textit{side-by-side evaluation} where raters compared multiple alternative report findings and ranked them based on their overall quality, and (2) \textit{independent evaluation} where raters assessed the quality of individual report findings.
Prior to performing the final evaluation, we iterated upon the instructions for the raters and calibrated their grades using a pilot set of 25 cases that were distinct from the evaluation set. Side-by-side evaluation was performed for all 246 cases, where each case was rated by a single radiologist randomly selected from a pool of four. For independent evaluation, each of the four radiologists independently annotated findings generated by three Med-PaLM M model variants (12B, 84B, and 562B) for every case in the evaluation set. Radiologists were blind to the source of the report findings for all evaluation tasks, and the reports were presented in a randomized order.

\paragraph{Side-by-side evaluation} The input to each side-by-side evaluation was a single chest X-ray, along with the ``indication'' section from the MIMIC-CXR study. Four alternative options for the ``findings'' section of the report were shown to raters as depicted in~\cref{fig-appendix:human-evaluation-ui-side-by-side}. The four alternative ``findings'' sections corresponded to the dataset reference report's findings, and findings generated by three Med-PaLM M model variants (12B, 84B, 562B).
Raters were asked to rank the four alternative findings based on their overall quality using their best clinical judgement.

\paragraph{Independent evaluation} For independent evaluation, raters were also presented with a single chest X-ray, along with the indication and reference report's findings from the MIMIC-CXR study (marked explicitly as such), but this time only a single findings paragraph generated by Med-PaLM M as shown in~\cref{fig-appendix:human-evaluation-ui-independent}. Raters were asked to assess the quality of the Med-PaLM M generated findings in the presence of the reference inputs provided and their own judgement of the chest X-ray image.
The rating schema proposed in~\citet{yu2022evaluating} served as inspiration for our evaluation task design.

First, raters assessed whether the quality and view of the provided image were sufficient to perform the evaluation task fully.
Next, they annotated all passages in the model-generated findings that they disagreed with (errors), and all missing parts (omissions).
Raters categorized each error passage by its type (no finding, incorrect finding location, incorrect severity, reference to non-existent view or prior study), assessed its clinical significance, and suggested alternative text to replace the selected passage. Likewise, for each omission, raters specified a passage that should have been included and determined if the omission had any clinical significance.

\begin{table}[ht]
\footnotesize
\centering

\caption{\textbf{Performance comparison on MultiMedBench.} We compare Med-PaLM M with specialist SOTA models and a generalist model (PaLM-E 84B) without biomedical domain finetuning. Across all tasks, datasets and metrics combination in MultiMedBench, we observe Med-PaLM M performance near or exceeding SOTA. Note that these results are achieved by Med-PaLM M with the same set of model weights without any task-specific customization.}
\label{tab:results-med-palm-m-best}
\begin{tabular}{@{}c@{\hspace{.05cm}}@{\hspace{.07cm}}c@{\hspace{.07cm}}ccccc}
\toprule
Task Type & Modality  &   Dataset   & Metric   & SOTA &\begin{tabular}[c]{@{}c@{}}PaLM-E \\ (84B)\end{tabular} &\begin{tabular}[c]{@{}c@{}}Med-PaLM M \\ (Best)\end{tabular} \\ \midrule

\multirow{3}{*}{Question Answering} & \multirow{3}{*}{Text}
& MedQA & Accuracy & \textbf{86.50\%}~\cite{singhal2023towards}       & 28.83\%         & 69.68\% \\
&& MedMCQA       & Accuracy & \textbf{72.30\%}~\cite{singhal2023towards}        & 33.35\%          & 62.59\% \\
&& PubMedQA      & Accuracy & \textbf{81.80\%}~\cite{singhal2023towards}        & 64.00\%          & 80.00\%    \\ 
\midrule
\multirow{3}{*}{Report Summarization} & \multirow{3}{*}{Radiology}
&\multirow{3}{*}{MIMIC-III} & ROUGE-L     & \textbf{38.70\%}~\cite{van2023radadapt}    & 3.30\%         & 32.03\% \\
&&& BLEU        & \textbf{16.20\%}~\cite{van2023radadapt}     & 0.34\%         & 15.36\% \\
&&& F1-RadGraph & \textbf{40.80\%}~\cite{van2023radadapt}           & 8.00\%         & 34.71\%\\
\midrule
\multirow{6}{*}{\begin{tabular}[c]{@{}c@{}}Visual \\Question Answering\end{tabular}} & \multirow{4}{*}{Radiology}
&\multirow{2}{*}{VQA-RAD}   & BLEU-1    & 71.03\%~\cite{bazi2023vision}   & 59.19\%          & \textbf{71.27\%} \\
                          
                           &&& F1          & N/A     & 38.67\%           & \textbf{62.06\%} \\ 
 &&\multirow{2}{*}{Slake-VQA} & BLEU-1      & 78.60\%~\cite{van2023open}   & 52.65\%        & \textbf{92.7\%}          \\
                          & && F1          & 78.10\%~\cite{van2023open}   & 24.53\%        & \textbf{89.28\%}         \\ 

& \multirow{2}{*}{Pathology} &\multirow{2}{*}{Path-VQA}  & BLEU-1   & 70.30\%~\cite{van2023open}  & 54.92\%          & \textbf{72.27\%} \\
                           
                           &&& F1          & 58.40\%~\cite{van2023open}    & 29.68\%          & \textbf{62.69\%} \\ 

\midrule
\multirow{9}{*}{Report Generation} &\multirow{9}{*}{Chest X-ray}
&\multirow{9}{*}{MIMIC-CXR} &Micro-F1-14 & 44.20\%~\cite{nicolson2022improving}  & 15.40\%        & \textbf{53.56\%} \\
&&&Macro-F1-14 & 30.70\%~\cite{nicolson2022improving}  & 10.11\%        & \textbf{39.83\%}  \\
&&&Micro-F1-5  & 56.70\%~\cite{miura2020improving}  & 5.51\%        & \textbf{57.88\%}  \\
&&&Macro-F1-5  & N/A     & 4.85\%        & \textbf{51.60\%}  \\
&&&F1-RadGraph         & 24.40\%~\cite{jeong2023multimodal}  & 11.66\%        & \textbf{26.71\%}          \\
&&&BLEU-1           & \textbf{39.48\%}~\cite{nicolson2022improving} & 19.86\%        & 32.31\% \\
&&&BLEU-4           & \textbf{13.30\%}~\cite{miura2020improving} & 4.60\%                & 11.50\%  \\
&&&ROUGE-L             & \textbf{29.60\%}~\cite{bannur2023learning} & 16.53\%         & 27.49\% \\
&&&CIDEr-D               & \textbf{49.50\%}~\cite{tanida2023interactive} & 3.50\%        & 26.17\%   \\

\midrule
\multirow{12}{*}{Image Classification} &\multirow{2}{*}{Chest X-ray}
&\multirow{2}{*}{\begin{tabular}[c]{@{}c@{}}MIMIC-CXR\\ (5 conditions)\end{tabular}} & Macro-AUC & \textbf{81.27\%}~\cite{rammuni2022effective}    & 51.48\%         & 79.09\% \\
&&& Macro-F1  & N/A       & 7.83\%        & \textbf{41.57\%} \\ 
&\multirow{2}{*}{Dermatology} 
&\multirow{2}{*}{PAD-UFES-20}  
& Macro-AUC & N/A & 63.37\%    & \textbf{97.27\%}      \\
&&& Macro-F1  & N/A       & 1.38\%          & \textbf{84.32\%}         \\ 

&\multirow{6}{*}{Mammography} 
&\multirow{2}{*}{VinDr-Mammo} & Macro-AUC & 64.50\%~\cite{wantlin2023benchmd}    & 51.49\%     & \textbf{71.76\%} \\
&&& Macro-F1  & N/A       & 16.06\%     & \textbf{35.70}\%             \\ 

&&\multirow{2}{*}{\begin{tabular}[c]{@{}c@{}}CBIS-DDSM\\ (mass)\end{tabular}}  &
  Macro-AUC & N/A & 47.75\%  & \textbf{73.31\%} \\
                                                   
&&& Macro-F1  & N/A       & 7.77\%        & \textbf{51.12\%} \\ 
&&\multirow{2}{*}{\begin{tabular}[c]{@{}c@{}}CBIS-DDSM\\ (calcification)\end{tabular}} &
  Macro-AUC & N/A & 40.67\% & \textbf{82.22\%}  \\
&&& Macro-F1  & \textbf{70.71\%}~\cite{panambur2022effect}     & 11.37\% & 67.86\%   \\ 
&\multirow{2}{*}{\begin{tabular}[c]{@{}c@{}}Genomics\\(Variant Calling)\end{tabular}}
&\multirow{2}{*}{\begin{tabular}[c]{@{}c@{}}PrecisionFDA\\(Truth Challenge V2)\end{tabular}}  & Indel-F1  & \textbf{99.40\%}~\cite{poplin2018deepvariant}   & 53.01\%          & 97.04\%        \\
                                 &  && SNP-F1    & \textbf{99.70\%}~\cite{poplin2018deepvariant}  & 52.84\%   & 99.35\%                \\                                  
\bottomrule

\end{tabular}
\end{table}

\begin{table}[ht]
\footnotesize
\centering
\caption{\textbf{Performance of Med-PaLM M on MultiMedBench across model scales.} We summarize the performance of Med-PaLM M across three model scale variants 12B, 84B, 562B. All models were finetuned and evaluated on the same set of tasks in MultiMedBench. We observe that scaling plays a key role in language-only tasks and multimodal tasks that require reasoning such as visual question answering. However, scaling has diminishing benefit for image classification and chest X-ray report generation task.}
\label{tab:results-med-palm-m-scaling}
\begin{tabular}{@{}c@{\hspace{.05cm}}@{\hspace{.07cm}}c@{\hspace{.07cm}}ccccc}
\toprule
Task Type & Modality  &   Dataset   & Metric     &\begin{tabular}[c]{@{}c@{}}Med-PaLM M \\ (12B)\end{tabular}  & \begin{tabular}[c]{@{}c@{}}Med-PaLM M \\ (84B)\end{tabular}  &  \begin{tabular}[c]{@{}c@{}}Med-PaLM M \\ (562B)\end{tabular}  \\ \midrule
\multirow{3}{*}{Question Answering} & \multirow{3}{*}{Text}
& MedQA  & Accuracy  & 29.22\%        & 46.11\%         & \textbf{69.68\%} \\
&& MedMCQA       & Accuracy  & 32.20\%         & 47.60\%          & \textbf{62.59\%} \\
&& PubMedQA      & Accuracy  & 48.60\%         & 71.40\%          & \textbf{80.00\%}    \\ 
\midrule
\multirow{3}{*}{Report Summarization} & \multirow{3}{*}{Radiology}
&\multirow{3}{*}{MIMIC-III} & ROUGE-L      & 29.45\%        & 31.47\%         & \textbf{32.03\%} \\
&&& BLEU          & 12.14\%        & \textbf{15.36}\%         & {15.21\%} \\
                          & && F1-RadGraph    & 31.43\%        & 33.96\%         & \textbf{34.71\%}\\
\midrule
\multirow{6}{*}{\begin{tabular}[c]{@{}c@{}}Visual \\Question Answering\end{tabular}} & \multirow{4}{*}{Radiology}
&\multirow{2}{*}{VQA-RAD}   & BLEU-1  & 64.02\%        & 69.38\%          & \textbf{71.27\%} \\
                          
                           &&& F1    & 50.66\%        & 59.90\%           & \textbf{62.06\%} \\ 
 &&\multirow{2}{*}{Slake-VQA} & BLEU-1    & 90.77\%       & \textbf{92.70\%}  & 91.64\%          \\
                          & && F1    & 86.22\%       & \textbf{89.28\%} & 87.50\%           \\ 

& \multirow{2}{*}{Pathology} &\multirow{2}{*}{Path-VQA}  & BLEU-1 & 68.97\%        & 70.16\%          & \textbf{72.27\%} \\
                           
                           &&& F1     & 57.24\%        & 59.51\%          & \textbf{62.69\%} \\ 

\midrule
\multirow{9}{*}{Report Generation} &\multirow{9}{*}{Chest X-ray}
&\multirow{9}{*}{MIMIC-CXR} &Micro-F1-14  & 51.41\%        & \textbf{53.56\%} & 51.60\% \\
&&&Macro-F1-14  & 37.31\%        & \textbf{39.83\%} & 37.81\% \\
&&&Micro-F1-5   & 56.54\%        & \textbf{57.88\%} & 56.28\% \\
&&&Macro-F1-5     & 50.57\%        & \textbf{51.60\%} & 49.86\% \\
&&&F1-RadGraph      & 25.20\%        & \textbf{26.71\%} & 26.06\%          \\
&&&BLEU-1            & 30.90\%        & \textbf{32.31\%} & 31.73\% \\
&&&BLEU-4           & 10.43\%        & 11.31\%          & \textbf{11.50\%}  \\
&&&ROUGE-L           & 26.16\%        & 27.29\% & \textbf{27.49\%} \\
&&&CIDEr-D           & 23.43\%        & \textbf{26.17\%} & 25.27\%          \\

\midrule
\multirow{12}{*}{Image Classification} &\multirow{2}{*}{Chest X-ray}
&\multirow{2}{*}{\begin{tabular}[c]{@{}c@{}}MIMIC-CXR\\ (5 conditions)\end{tabular}} & Macro-AUC    & 76.67\%          & 78.35\%          & \textbf{79.09\%} \\
                                & & & Macro-F1      & 38.33\%          & 36.83\%          & \textbf{41.57\%} \\ 
&\multirow{2}{*}{Dermatology} 
&\multirow{2}{*}{PAD-UFES-20}  
& Macro-AUC & 95.57\%          & \textbf{97.27\%} & 96.08\%          \\
                        & && Macro-F1        & 78.42\%          & \textbf{84.32\%} & 77.03\%          \\ 

&\multirow{6}{*}{Mammography} 
&\multirow{2}{*}{VinDr-Mammo} & Macro-AUC     & 66.29\%          & \textbf{71.76\%} & 71.42\%          \\
&&& Macro-F1        & 29.81\%    & \textbf{35.70}\%           & 33.90\%           \\ 

&&\multirow{2}{*}{\begin{tabular}[c]{@{}c@{}}CBIS-DDSM\\ (mass)\end{tabular}}  &
  Macro-AUC  &
  70.11\% &
  73.09\% &
  \textbf{73.31\%} \\
                                                   
                                 &                 &          & Macro-F1        & 47.23\%          & 49.98\%          & \textbf{51.12\%} \\ 
&&\multirow{2}{*}{\begin{tabular}[c]{@{}c@{}}CBIS-DDSM\\ (calcification)\end{tabular}} &
  Macro-AUC  & 81.40\% &
  \textbf{82.22\%} &
  80.90\% \\
                                 &                  &         & Macro-F1        & \textbf{67.86}\%          & 63.81\% & 63.03\%          \\ 
&\multirow{2}{*}{Genomics} 
&\multirow{2}{*}{Variant Calling}  & Indel-F1    & 96.42\%          & \textbf{97.04\%} & 95.46\%          \\
                                 &  && SNP-F1      & \textbf{99.35\%} & 99.32\%          & 99.16\%          \\                                  
\bottomrule

\end{tabular}
\end{table}

\section{Results}
\label{sec:results}

Here we present results across the three different evaluation setups introduced in~\cref{sec:evaluation}.

\subsection{Med-PaLM M performs near or exceeding SOTA on all MultiMedBench tasks}

\paragraph{Med-PaLM M performance versus baselines} We compared Med-PaLM M with two baselines:
\begin{itemize}
    \item prior SOTA specialist models for each of the MultiMedBench tasks
    \item a baseline generalist model (PaLM-E 84B) without any biomedical domain finetuning. We used this model size variant (and not PaLM-E 562B) due to compute constraints.
\end{itemize}

Results are summarized in~\cref{tab:results-med-palm-m-best}. Across MultiMedBench tasks, Med-PaLM M's best result (across three model sizes) exceeded prior SOTA results on 5 out of 12 tasks (for two tasks, we were unable to find a prior SOTA comparable to our setup) while being competitive on the rest. Notably, these results were achieved with a generalist model using the same set of model weights without any task-specific architecture customization or optimization. 

On medical question answering tasks, we compared against the SOTA Med-PaLM 2 results~\cite{singhal2023towards} and observed higher performance of Med-PaLM 2. However, when compared to the baseline PaLM model on which Med-PaLM M was built, Med-PaLM M outperformed the previous best PaLM results~\cite{singhal2022large} by a large margin in the same few-shot setting on all three question answering datasets.

Further, when compared to PaLM-E 84B as a generalist baseline without biomedical domain finetuning, Med-PaLM M exhibited performance improvements on all 14 tasks often by a significant margin, demonstrating the importance of domain adaptation. Taken together, these results illustrate the strong capabilities of Med-PaLM M as a generalist biomedical AI model. We further describe the results in detail for each of the individual tasks in~\cref{appendix:multimedbench_detailed_performance}.

\paragraph{Med-PaLM M performance across model scales} We summarize Med-PaLM M performance across model scales (12B, 84B, and 562B) in~\cref{tab:results-med-palm-m-scaling}. The key observations are:
\begin{itemize}
    \item \textbf{Language reasoning tasks benefit from scale} For tasks that require language understanding and reasoning such as medical question answering, medical visual question answering and radiology report summarization, we see significant improvements as we scale up the model from 12B to 562B.
    \item \textbf{Multimodal tasks bottlenecked by vision encoder performance} For tasks such as mammography or dermatology image classification, where nuanced visual understanding is required but minimal language reasoning is needed (outputs are classification label tokens only), the performance improved from Med-PaLM M 12B to Med-PaLM 84B but plateaued for the 562B model, possibly because the vision encoder is not further scaled in that step (both the Med-PaLM M 84B and 562B models use the same 22B ViT as the vision encoder), thereby acting as a bottleneck to observing a scaling benefit. We note the possibility of additional confounders here such as the input image resolution.  
\end{itemize}

The scaling results on the chest X-ray report generation task are interesting (\cref{tab:results-med-palm-m-scaling}). While on the surface, the task seems to require complex language understanding and reasoning capabilities and would thus benefit from scaling the language model, we find the Med-PaLM M 84B model to be roughly on-par or slightly exceeding the 562B model on a majority of metrics, which may simply be due to fewer training steps used for the larger model.
Another possibility for the diminishing return of increasing the size of language model is likely that the output space for chest X-ray report generation in the MIMIC-CXR dataset is fairly confined to a set of template sentences and limited number of conditions. This insight has motivated the use of retrieval based approaches as opposed to a fully generative approach for the chest X-ray report generation task on this dataset ~\cite{ye2022retrieval, endo2021retrieval}. Additionally, the larger 562B model has a tendency towards verbosity rather than the comparative brevity of the 84B model, and without further preference alignment in training, this may impact its metrics.

\subsection{Med-PaLM M demonstrates zero-shot generalization to novel medical tasks and concepts}

Training a generalist biomedical AI system with language as a common grounding across different tasks allows the system to tackle new tasks by combining the knowledge it has learned for other tasks (i.e. combinatorial generalization).
We highlight preliminary evidence which suggests Med-PaLM M can generalize to novel medical concepts and unseen tasks in a zero-shot fashion. We further observe zero-shot multimodal reasoning as an emergent capability~\cite{wei2022emergent} of Med-PaLM M. Finally, we demonstrate benefits from positive task transfer as a result of the model's multi-task, multimodal training.

\subsubsection{Evidence of generalization to novel medical concepts}
We probed the zero-shot generalization capability of Med-PaLM M for an unseen medical concept by evaluating its ability to detect tuberculosis (TB) abnormality from chest X-ray images in the Montgomery County (MC) dataset. As shown in~\cref{tab:results-tb-classification}, Med-PaLM M performed competitively compared to SOTA results obtained by a specialized ensemble model optimized for this dataset~\cite{oloko2021ensemble}. We observed similar performance across three model variants, consistent with findings on other medical image classification tasks in MultiMedBench. Given the classification task was set up as an open-ended question answering task, we did not report the AUC metric which requires the normalized predicted probability of each possible class. 

\begin{table}[ht]
\small
\centering
\caption{\textbf{Zero-shot classification performance of Med-PaLM M on the tuberculosis (TB) detection task.} Med-PaLM M performs competitively to the SOTA model~\cite{oloko2021ensemble} finetuned on the Montgomery County TB dataset using model ensemble. Notably, Med-PaLM M achieves this result with a simple task prompt consisting of a single text-only exemplar (without task-specific image and hence zero-shot), in contrast to the specialist model that requires training on all the samples in the dataset.}
\label{tab:results-tb-classification}
\begin{tabular}{ccc}
\toprule

  Model & \# Training samples & Accuracy\\ \midrule
  SOTA~\cite{oloko2021ensemble}  & \textbf{138} &   \textbf{92.60\%} \\
  Med-PaLM M (12B) & 0 & 86.96\% \\
  Med-PaLM M (84B) & 0 & 82.60\% \\  
  Med-PaLM M (562B) & 0 & 87.68\% \\
  \bottomrule
\end{tabular}
\end{table}

\subsubsection {Evidence of emergent zero-shot multimodal medical reasoning}
\begin{figure*}[t]
\small
    \centering
    \includegraphics[width=\textwidth]{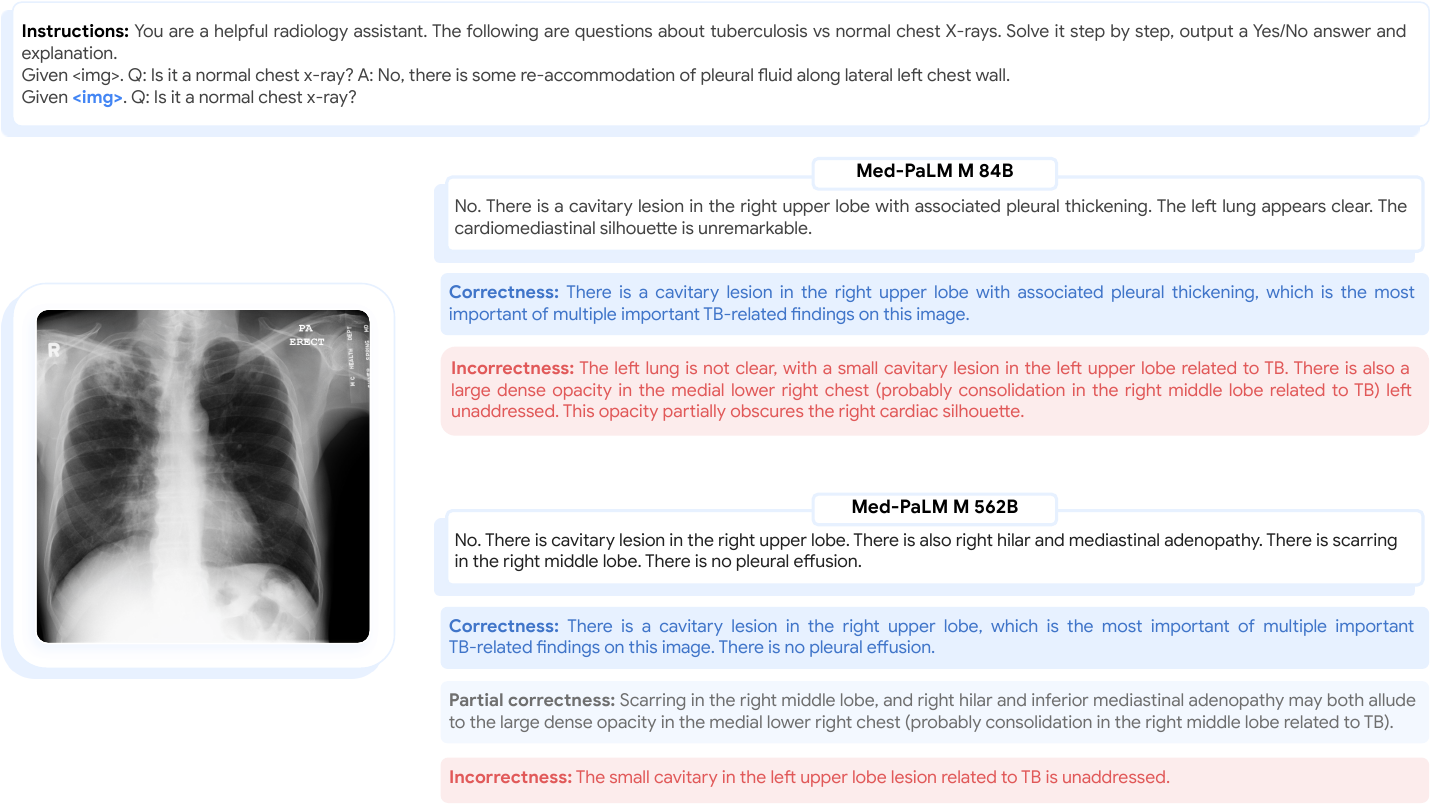} 
    \vspace{6pt}
    \caption{\textbf{Evidence of emergent zero-shot multimodal medical reasoning with Med-PaLM M.} Large Med-PaLM M models exhibit zero-shot CoT reasoning capability in identifying and describing tuberculosis related findings in chest X-ray images. The model is prompted with task-specific instructions and a text-only exemplar (without the corresponding image) to generate a report describing findings in the given X-ray image. Model predictions from Med-PaLM M 84B and 562B are shown together with the annotations from an expert radiologist. Both models correctly localized the major TB related cavitory lesion in the right upper lobe. However, both models did not address the small cavitory lesion in left upper lobe (Med-PaLM M 562B was considered better than Med-PaLM M 64B in this example as it also alluded to the opacity in the right middle lobe and did not make the incorrect statement of left lung being clear). Notably, Med-PaLM M 12B failed to generate a coherent report, indicating the importance of scaling for zero-shot COT reasoning.}
    \vspace{-0pt}
    \label{fig:tb-zero-shot-cot}
\end{figure*}

We also qualitatively explored the zero-shot chain-of-thought (CoT) capability of Med-PaLM M on the MC TB dataset. In contrast to the classification setup, we prompted the model with a text-only exemplar to generate a report describing the findings in a given image in addition to a yes/no classification prediction.  In~\cref{fig:tb-zero-shot-cot}, we present qualitative examples of zero-shot CoT reasoning from the Med-PaLM M 84B and 562B variants. In particular, both Med-PaLM M variants were able to identify the major TB related lesion in the correct location. However, according to expert radiologist review, there are still some omissions of findings and errors in the model generated report, suggesting room for improvement. It is noteworthy that Med-PaLM M 12B failed to generate a coherent visually conditioned response, which indicates that scaling of the language model plays a key role in the zero-shot CoT multimodal reasoning capability (i.e. this might be an emergent capability~\cite{wei2022emergent}).

\subsubsection{Evidence of generalization to novel tasks}
Although Med-PaLM M was only trained with single-view chest X-ray image inputs, we observed the capability of the model to generalize to a novel task setup with multi-view visual inputs. Specifically, on a subset of studies from MIMIC-CXR where each report is accompanied with both a frontal and a lateral view X-ray image. we observe that Med-PaLM M is able to attain zero-shot performance comparable to the single-view report generation task as detailed in~\cref{tab:results-report-gen-ood}.  This ability is promising given medical imaging studies often benefit from the interpretation of prior historical studies in addition to the current instance for optimal performance.

\begin{table}[ht]
\small
\centering
\caption{\textbf{Zero-shot generalization to two-view chest X-ray report generation.} Med-PaLM M performance remains competitive on a novel two-view report generation task setup despite having not been trained with two visual inputs before. Med-PaLM M achieves SOTA results on clinical efficacy metrics for the two view report generation task.}
\label{tab:results-report-gen-ood}
\begin{tabular}{ccccc}
\toprule
Metric    & SOTA    & Med-PaLM M (12B)   & Med-PaLM M (84B)  & Med-PaLM M (562B) \\ \midrule
Micro-F1-14  & 44.20\% & 49.80\% & \textbf{50.54\%} & 48.85\%          \\
Macro-F1-14  &30.70\% & 37.69\% & \textbf{37.78\%} & 37.29\%          \\
Micro-F1-5   & 56.70\% & 54.49\% & \textbf{56.37\%} & 54.36\%          \\
Macro-F1-5   & N/A  & 48.33\% & \textbf{51.23\%} & 48.49\%          \\
F1-RadGraph  & 24.40\% & 26.73\% & \textbf{28.30\%} & 27.28\%  \\
BLEU-1       & \textbf{39.48\%} & 33.31\% & 34.58\% & 33.83\%          \\
BLEU-4       & \textbf{13.30\%} & 11.51\%          & 12.44\% & 12.47\%          \\
ROUGE-L      & \textbf{29.60\%} & 27.84\% & 28.71\% & 28.49\%          \\
CIDEr-D       & \textbf{49.50\%} & 27.58\% & 29.80\%  & 29.80\%           \\ \bottomrule
\end{tabular}
\end{table}

\subsubsection{Evidence of positive task transfer}
To demonstrate the positive task transfer arising from joint training across modalities and tasks, we performed an ablation study where we trained a Med-PaLM M 84B variant by excluding the MIMIC-CXR classification task from the task mixture and compared this model variant against Med-PaLM M 84B trained on the full MultiMedBench mixture.
As seen in~\cref{tab:results-cxr-classification-positive-transfer}, we observed that the model trained jointly on both report generation and classification has higher performance across the board on all report generation metrics.
We also observe that the model trained only on chest X-ray report generation can generalize to abnormality detection in a zero-shot fashion with compelling performance, as evidenced by a higher macro-F1 score. This is another example of generalization to a novel task setting where the model learns to differentiate between types of abnormalities from training on the more complex report generation task.

\begin{table}[ht]
\small
\centering
\caption{\textbf{Positive task transfer between CXR report generation and abnormality classification.} We observe positive transfer as a result of multi-task training with Med-PaLM M model trained jointly on both chest X-ray report generation and classification tasks. It exhibits higher performance on report generation metrics compared to a Med-PaLM M model trained without chest X-ray report classification. We also observe that training on the chest X-ray report generation task alone enables Med-PaLM M to generalize to abnormality detection in a zero-shot fashion.}
\label{tab:results-cxr-classification-positive-transfer}
\begin{tabular}{cccc}
\toprule
Dataset &
  Metric &
  Med-PaLM M (84B) &
  \begin{tabular}[c]{@{}c@{}}Med-PaLM M (84B)\\ No CXR classification\end{tabular} \\ \midrule
\multirow{9}{*}{MIMIC-CXR} &
  Micro-F1-14 & \textbf{53.56\%} & 52.94\% \\
 & Macro-F1-14 & \textbf{39.83\%} & 38.92\% \\
 & Micro-F1-5  & \textbf{57.88\%} & 57.58\% \\
 & Macro-F1-5  & \textbf{51.60\%} & 51.32\% \\
 & F1-RadGraph         & \textbf{26.71\%} & 26.08\% \\
 & BLEU-1            & \textbf{32.31\%} & 31.72\% \\
 & BLEU-4            & \textbf{11.31\%} & 10.87\%   \\
 & ROUGE-L             & \textbf{27.29\%} & 26.67\% \\
 & CIDEr-D               & \textbf{26.17\%} & 25.17\% \\
\midrule
\multirow{2}{*}{\begin{tabular}[c]{@{}c@{}}MIMIC-CXR\\ (5 conditions)\end{tabular}}  &
  Macro-AUC &\textbf{78.35\%} & 73.88\% \\
 & Macro-F1            & 36.83\% & \textbf{43.97}\%  \\ \bottomrule
\end{tabular}
\end{table}

\subsection{Med-PaLM M performs encouragingly on radiology report generation across model scales}

To further understand the clinical applicability of Med-PaLM M, we conducted radiologist evaluations of model-generated chest X-ray reports (and reference human baselines). Under this evaluation framework, we observe encouraging quality of Med-PaLM M generated reports across model scales as detailed below.

\subsubsection{Side-by-side evaluation}

In a side-by-side evaluation, four clinician raters ranked the quality of four radiology reports, comparing the radiologist-provided reference report from the MIMIC-CXR dataset with reports generated by different Med-PaLM M model scales (12B, 84B, and 562B).

\begin{figure}[t]
\centering
     \begin{subfigure}[t]{0.405\textwidth}
     \centering
         \includegraphics[width=\textwidth]{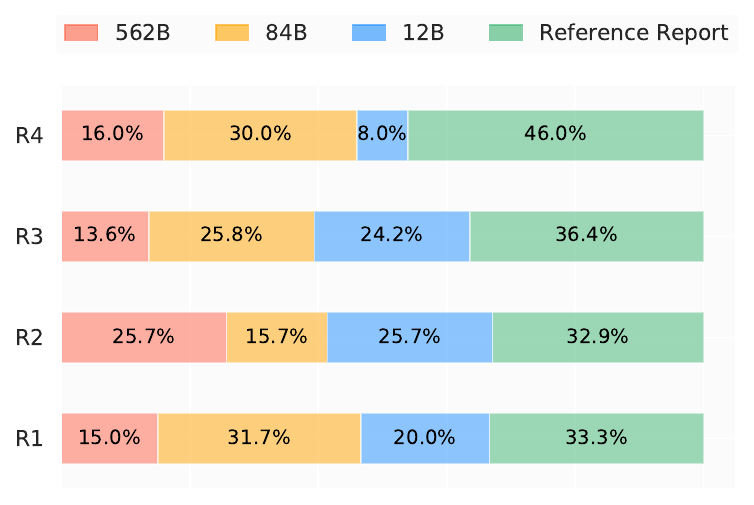}
         \caption{Best-ranked report in four-way comparison}
         \label{fig:ranking-best-of-four}
     \end{subfigure}
     \hfill
     \begin{subfigure}[t]{0.545\textwidth}
     \centering
         \includegraphics[width=\textwidth]{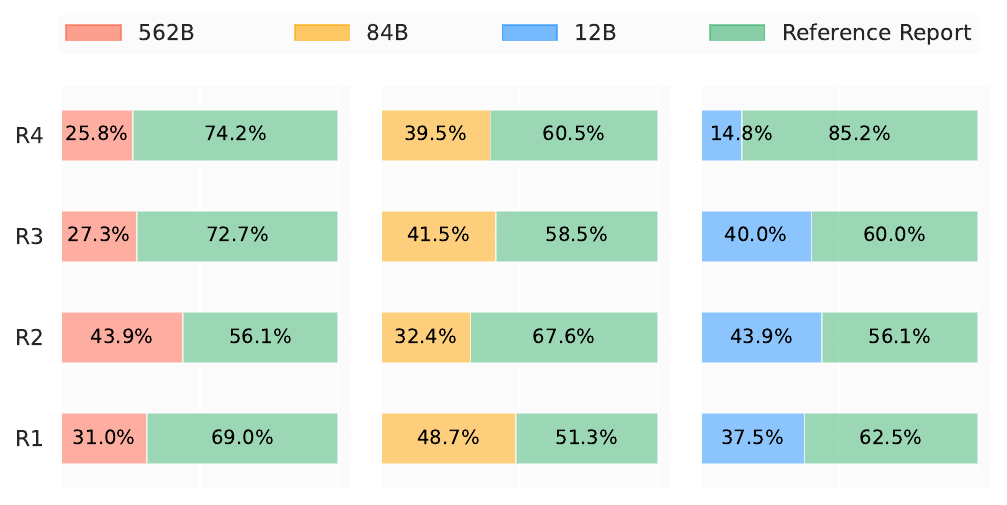}
         \caption{Pairwise preference of each model scale compared to reference report}
         \label{fig:ranking-pairwise-with-reference}
     \end{subfigure} 
     \centering
     \caption{\textbf{Side-by-side human evaluation.} Four clinician raters ranked the quality of four radiology reports in a side-by-side evaluation, comparing the radiologist-provided reference report from MIMIC-CXR with reports generated by different Med-PaLM M model scale variants (12B, 84B, 562B).}
     \label{fig:ranking-full}
\end{figure}

\cref{fig:ranking-best-of-four} summarizes how often each rater ranked a report generated by one of the three Med-PaLM M variants or the reference report as the best among four candidate reports.
Averaged over all four raters, the radiologist-provided reference report was ranked best in 37.14\% of cases, followed by Med-PaLM M (84B) which was ranked best in 25.78\% of cases, and the other two model scales, 12B and 562B, which were ranked best in 19.49\% and 17.59\% of cases respectively.

To enable a direct comparison of reports generated by each Med-PaLM M model scale to the radiologist-provided reference report, we derived pairwise preferences from the four-way ranking and provided a breakdown for each rater and model scale in~\cref{fig:ranking-pairwise-with-reference}.
Averaged over all four raters, Med-PaLM M 84B was preferred over the reference report in 40.50\% of cases, followed by the other two model scales, 12B and 562B, which were preferred over the reference report in 34.05\% and 32.00\% of cases, respectively.

\subsubsection{Independent evaluation}

\begin{figure*}[t]
\small
    \centering
    \includegraphics[width=1.0\textwidth]{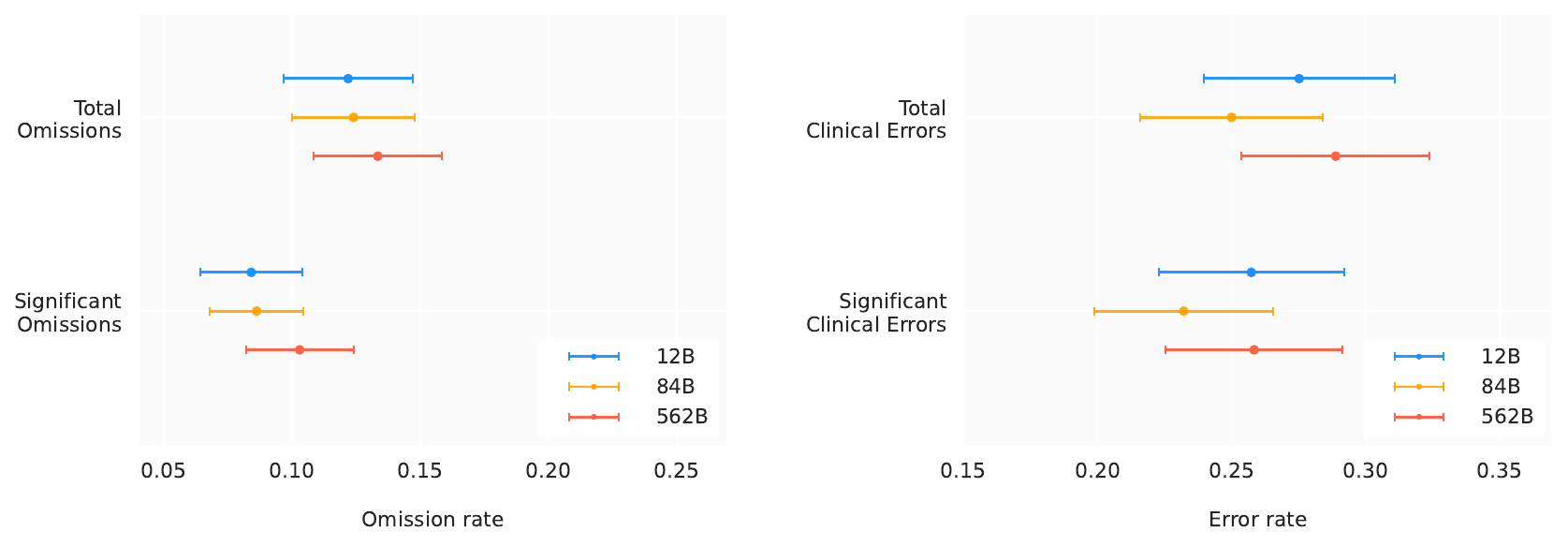}
    \vspace{0pt} 
    \caption{\textbf{Independent human evaluation.} Rates of omissions and clinical errors identified by clinician raters in radiology reports generated by Med-PaLM M. Clinical errors are those related to the presence, location or severity of a clinical finding.}
    \vspace{-0pt} 
    \label{fig:expert-error}
\end{figure*}

We report the rates of omissions and errors radiologists identified in findings paragraphs generated by Med-PaLM M.~\cref{fig:expert-error} provides breakdowns by model scales (12B, 84B, 562B). We observed different trends for omissions and errors. For omissions, we observed the lowest rate of 0.12 (95\% CI, 0.10 - 0.15) omissions per report on average for both the Med-PaLM M 12B and 84B models, followed by 0.13 (95\% CI, 0.11 - 0.16) for the 562B model.

In contrast, we measured the lowest mean error rate of 0.25 (95\% CI, 0.22 - 0.28) for Med-PaLM M 84B, followed by 0.28 (95\% CI, 0.24 - 0.31) for Med-PaLM M 12B and 0.29 (95\% CI, 0.25 - 0.32) for the 562B model. Notably, this error rate is comparable to those reported for human radiologists baselines on the MIMIC-CXR dataset in a prior study~\cite{jeong2023multimodal}. 

It is important to mention that our analysis is limited to errors of clinical relevance, ensuring a specific focus on clinical interpretation. This includes those errors related to the presence, location or severity of a clinical finding. Example of non-clinical errors are passages referring to views or prior studies not present, which stem from training artifacts.

These trends across model scales were identical for the subset of omissions and errors that were marked as significant by radiologist raters. We refer the reader to~\cref{tab-appendix:human-evaluation-results} for an overview of error and omission rates, including non-clinical errors.

\begin{figure*}[ht]
\small
    \centering
    \includegraphics[width=1\textwidth]{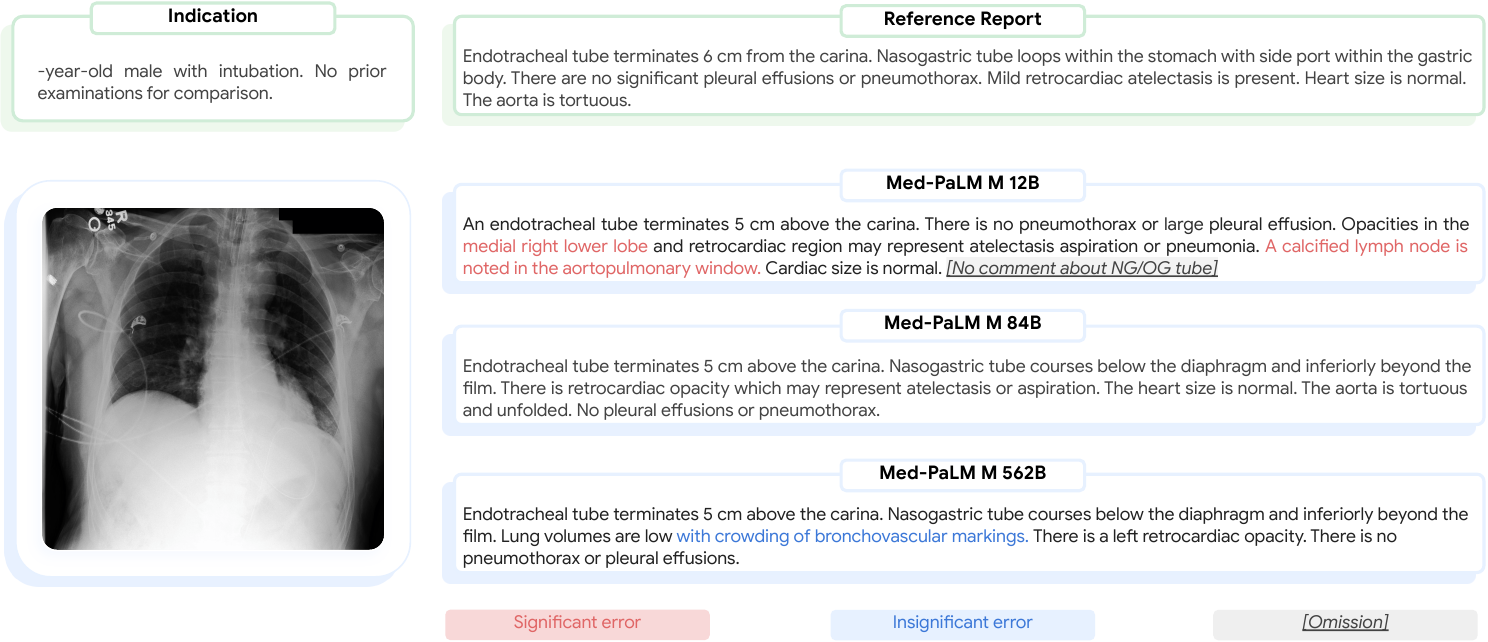} 
    \vspace{6pt}
    \caption{\textbf{Qualitative example of reference and Med-PaLM M generated chest X-ray reports.} We present a qualitative example of chest X-ray reports generated by Med-PaLM M across model scales along with the target reference report. In this example, a panel of radiologists adjudicated the Med-PaLM M 12B report to have two clinically significant errors and one omission, the Med-PaLM M 84B report to have zero errors and zero omissions, and the Med-PaLM M 562B report to have one clinically insignificant error and no omissions.}
    \vspace{-0pt}
    \label{fig:cxr-report-example}
\end{figure*}

In~\cref{fig:cxr-report-example}, we illustrate a qualitative example of chest X-ray reports generated by Med-PaLM M across three model sizes along with the target reference report. For this example, our panel of radiologists judged the Med-PaLM M 12B report to have two clinically significant errors and one omission, the Med-PaLM M 84B report to have zero errors and zero omissions, and the Med-PaLM M 562B report to have one clinically insignificant errors and no omissions.

\section{Discussion}

To the best of our knowledge, Med-PaLM M is the first demonstration of a generalist biomedical AI system that can interpret a wide range of medical modalities, perform competently (including near or exceeding prior SOTA) on a diverse array of tasks, and generalize to unseen biomedical concepts and tasks. This potentially opens up new possibilities in applications spanning scientific discovery to care delivery. We elaborate on the implications of this development as well as the challenges and limitations below.

\paragraph{Lack of benchmarks a key bottleneck for the development of generalist biomedical AI} AI progress to date has largely been catalyzed by the development of high quality benchmarks. While there exists several single-task biomedical AI datasets, there have been limited attempts to unify them and create benchmarks for the development of generalist biomedical AI systems. Our curation of MultiMedBench is a step towards addressing this unmet need. However, the benchmark has several important limitations including limited size of the individual datasets (a cumulative size of \~1 million samples) and limited modality and task diversity (e.g., lacking life sciences such as transcriptomics and proteomics). Another key barrier to developing models for use across an even wider variety of biomedical data types is the lack of large scale multimodal datasets, which would permit joint learning and alignment of the modality-specific encoders with the decoder.

\paragraph{Importance of medical finetuning and specialization} PaLM-E is a highly capable generalist AI model as evidenced by its SOTA performance on a wide range of vision-language and embodied robotics tasks. Yet, its out-of-the-box performance on MultiMedBench was poor and Med-PaLM M outperforms it by a wide margin across model scales.  This result suggests that finetuning with domain-specific biomedical data is critical to achieving good performance on biomedical tasks, perhaps due to the distribution shift presented by the domain overall compared to the plethora of non-medical tasks and modalities.

\paragraph{Scaling multimodal AI models is challenging} In the language domain, scaling the model has led to leapfrog improvements in performance and emergent capabilities. However, our preliminary experiments suggest this is likely more challenging for multimodal generalist models in the biomedical task domain due to the medical data scarcity. Given the wide array of modalities and tasks such generalist models are expected to understand and tackle, it is crucial that the encoders for such diverse modalities are scaled jointly with the language model. Otherwise, for tasks that require interpretation of data from a combination of modalities, the performance will end up being bottlenecked by the weakest encoder. We see evidence of this in medical image classification tasks such as mammography and dermatology where scaling the language model component has little effect on the task performance as the potential key bottleneck is the vision encoder. It is possible that the small volume of medical data in MultiMedBench is not be sufficient to effectively adapt a ViT pretrained on natural images to the medical domain, thereby limiting the benefits of model scaling. As such, our study only provides some initial insights on the effect of model scaling on biomedical task performance. Future research is needed to fully understand the effect of model scaling by teasing apart the scaling effect of the language model from that of modality-specific encoders, with sufficient amounts of biomedical data.

\paragraph{Technical considerations for generalist biomedical AI} Med-PaLM M builds on state-of-the-art vision and language components such as ViT and PaLM. Yet, putting them together requires careful considerations around token lengths allocated to visual encoder outputs, total context length of the model, sampling strategies, training data mixtures and so forth. Further, simple, yet important techniques such as the use of one-shot training with dummy image tokens make an important difference in the quality and compute efficiency of the final model. With increasing generality of the AI system, the number of details requiring careful consideration tends to increase as well. We also note that Med-PaLM M architecture as setup currently is not optimal for few-shot in-context learning.

\paragraph{Progress in AI for radiology report generation} Our evaluation by radiologists of Med-PaLM M generated radiology reports suggests encouraging performance of the model on a challenging multimodal task. In up to 40.50\% of the cases, a Med-PaLM M generated report was preferred over the human-generated reference report. Further, the average number of clinically significant errors within the model responses is comparable to those reported for human-generated reports in prior studies~\cite{jeong2023multimodal} on the same dataset. These promising results underpin rapid development in the task of automatic radiology report generation and suggest the potential for clinical utility in the future.

\paragraph{Generalist agents are not the only approach to multimodal biomedical AI} While generalist biomedical AI systems offer exciting possibilities~\cite{moor2023foundation}, there are other approaches to developing multimodal biomedical AI systems that might be more applicable depending on data availability, pretrained models, compute and application scenarios. These include leveraging frozen encoders with adapter layers~\cite{zhang2023llama} to glue together a multimodal biomedical AI system or developing LLMs that can interface with specialist biomedical encoders or task-specific agents through tool use~\cite{schick2023toolformer}. 

\paragraph{Considerations for real-world applications of generalist biomedical AI} While the development of generally capable biomedical AI systems is exciting, for such systems to be useful in practice or opening the door to new applications, they need to match or exceed specialized, single-task models or otherwise reach clinically applicable levels of performance. While beyond the scope of this work, the progress here necessitates careful considerations of safety and equity in the development and validation of such systems.

\section{Perspective on Generalist Biomedical AI}
\label{sec:perspective}
Reaching near or above SOTA on a diverse range of biomedical tasks with a single set of model weights is a noteworthy milestone for the development of generalist biomedical AI systems. While human clinicians can train for ``general practice''~\cite{marshall2022future}, helpful subspecialty-specific expertise is often found in different experts~\cite{blank2014referral}, to whom non-specialist clinicians may refer for specialist opinions in the course of care. It is also commonplace for multiple physician specialities to work together in care delivery. We envisage a similar future for biomedical AI where generalist and specialist AI systems interact and collaborate together with expert clinicians and researchers in a tight feedback loop to tackle grand challenges in biomedicine. 

Our finding, of a single generalist biomedical AI that reaches compelling performance across disparate tasks and contexts, hints at new frontiers for impact in applications. This includes the potential for near zero-shot insight in new domains, as a tool for discovery integrating insights from distinct areas of biomedicine, and as a common point of assistance providing access to expertise from many different fields.

\section{Conclusion}
\label{sec:conclusion}
Medicine is a multidisciplinary endeavour. Generalist biomedical AI systems that effectively assimilate and encode multimodal medical data at scale and rapidly adapt to new clinical contexts are likely to be the foundation of next generation learning health systems and make healthcare more accessible, efficient, equitable and humane. While further development and rigorous validation is needed, we believe Med-PaLM M represents an important step towards the development of such generalist biomedical AI.

\vspace{12pt}
\subsubsection*{Acknowledgments}
This project was an extensive collaboration between many teams at Google Research and Google DeepMind. We thank Andrew Sellergren, Yuan Liu, Michael Howell, Julie Wang, Sho Kannan, Christine Kingsley, Roy Lee, Naama Hammel, Jay Hartford, Preeti Singh, Kavita Kulkarni, Gavriel Goidel, Anil Palepu, Si Wai Man, Amy Wang, Sami Lachgar, Lauren Winer, Maggie Shiels, Annisah Um'rani, John Guilyard, Shravya Shetty and Evan Rapoport for their valuable insights and feedback during our research. We are also grateful to Karen DeSalvo, Zoubin Ghahramani, James Manyika, and Jeff Dean for their support during the course of this project.

\subsubsection*{Data Availability}
The benchmark used for training and evaluation in this study, MultiMedBench, comprises de-identified datasets that are all open source. We present an overview of datasets in \cref{tab:multimedbench-overview}.

\subsubsection*{Code Availability} We will not be able to open source the large language models (LLMs) used in this study. We have provided comprehensive details regarding our underlying methodology and build on previously detailed models \cite{chowdhery2022palm,driess2023palme}, so that similar approaches can be tried with other classes of LLMs.

\newpage
\setlength\bibitemsep{3pt}
\printbibliography
\balance
\clearpage

\newpage
\clearpage
\onecolumn

\renewcommand{\thesection}{A.\arabic{section}}
\renewcommand{\thefigure}{A.\arabic{figure}}
\renewcommand{\thetable}{A.\arabic{table}} 
\renewcommand{\theequation}{A.\arabic{equation}} 
\renewcommand{\theHsection}{A\arabic{section}}

\setcounter{section}{0}
\setcounter{figure}{0}
\setcounter{table}{0}
\setcounter{equation}{0}

\noindent \textbf{\LARGE{Appendix}}\\
\normalfont

In the following sections, we report additional experiments and detailed analysis to further illustrate the performance of our proposed generalist model, Med-PaLM M.

We provide details on:

\begin{itemize}[leftmargin=1.5em,rightmargin=0em]

\item  Datasets and tasks in MultiMedBench

\item  Med-PaLM M training procedure 
\item  Interpretations of Med-PaLM M performance by task type:
\begin{itemize}
    \item Performance analysis on language-only medical question answering
    \item Performance analysis on radiology report summarization
    \item Performance analysis on medical image classification tasks
    \item Performance analysis on medical visual question answering
    \item Performance analysis on chest X-ray report generation
\end{itemize}
\item  Human evaluation of model-generated chest X-ray reports
\item  Examples from MultiMedBench tasks

\end{itemize}

\section{MultiMedBench}
\label{appendix:multimedbench-details}

\begin{figure}[t]
\centering
     \includegraphics[width=0.9\textwidth]{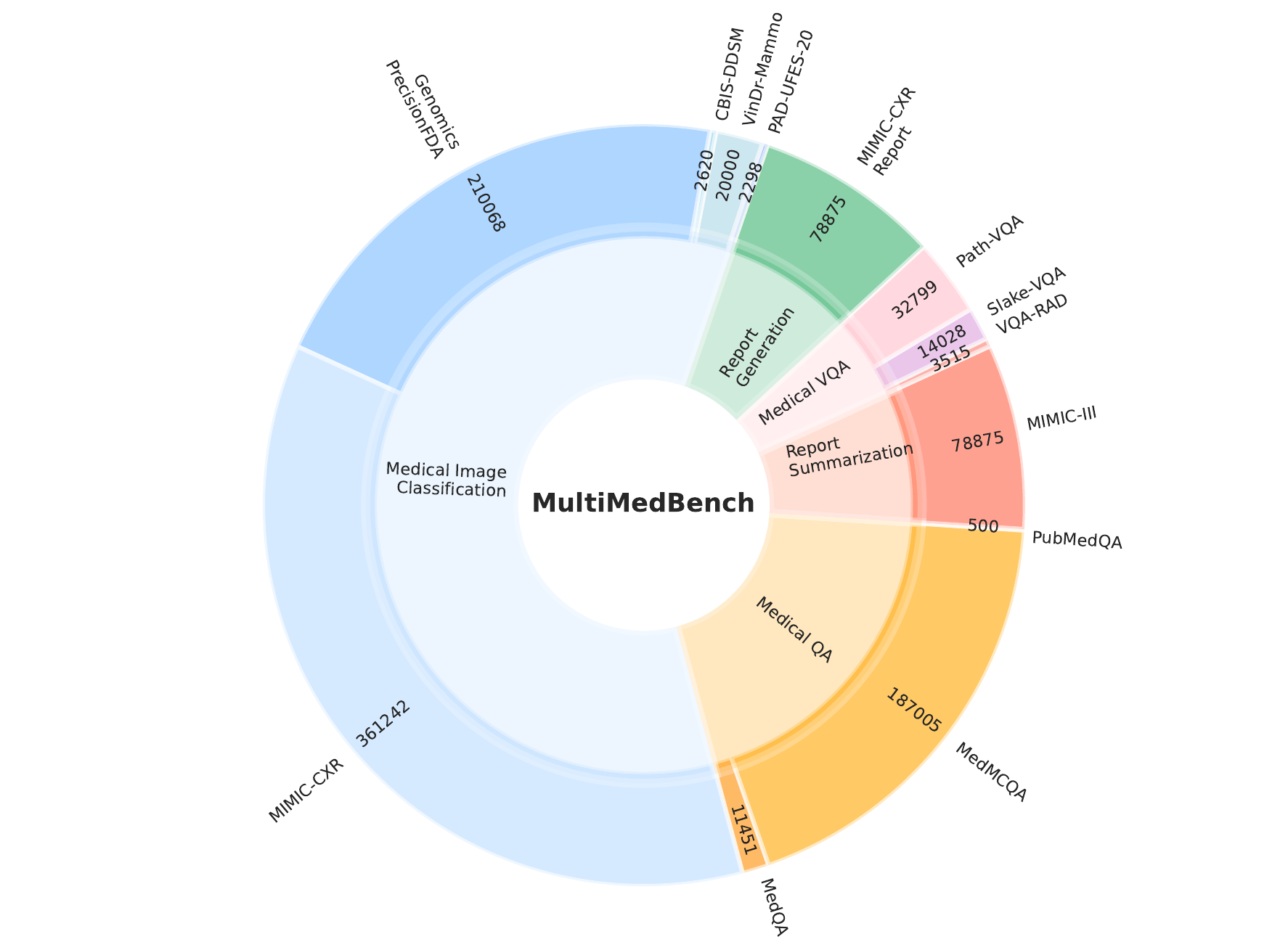}
     \vspace{6pt}
     \caption{\textbf{MultiMedBench overview.} MultiMedBench is a benchmark that covers 14 different biomedical tasks, including question answering, visual question answering, image classification, radiology report generation and summarization, and genomic variant calling. MultiMedBench comprises more than 1 million data samples from a diverse range of medical images, radiology reports, medical question answers, and visual question answering pairs.}
     \label{fig:datasets-overiew}
\end{figure}

In this section, we offer a comprehensive overview of \emph{MultiMedBench}, including a detailed description of the datasets, data preprocessing, and task setups. Figure~\ref{fig:datasets-overiew} summarizes MultiMedBench over its various biomedical tasks.

\subsection{Language-only datasets}
\paragraph{MultiMedQA}
We used three of the multiple-choice medical question-answering datasets from
MultiMedQA~\cite{singhal2022large}: the MedQA~\cite{jin2021disease}, MedMCQA~\cite{pal2022medmcqa}, and PubMedQA~\cite{jin2019pubmedqa} datasets for training and evaluation of Med-PaLM M. These question answering tasks are language-only and do not require the interpretation of additional modalities. The training set consists of 10,178 questions from MedQA and 182,822 questions from MedMCQA. The test set comprises 1,273 questions from MedQA, 4,183 questions from MedMCQA, and 500 questions from PubMedQA. Note that PubMedQA was not included in the training data mixture and only used for evaluation.

\paragraph{MIMIC-III} is a large publicly-available medical database that contains medical records of patients admitted to intensive care units~\cite{johnson2016mimic}. It contains 79,790 radiology reports across two imaging modalities (CT and MRI) and seven anatomic regions (head, abdomen, chest, head, neck, sinus, spine, pelvis). A total of 78,875 reports were chosen based on criteria such as the length of the report. We used the radiology report summarization dataset from~\cite{van2023radadapt}, which comprises six most common modality/anatomy pairs for training and evaluation: CT head, CT abdomen, CT chest, MRI head, CT spine, and CT neck. To evaluate out-of-distribution (OOD) performance we used five less common modality/anatomy pairs: MRI spine, CT sinus, MRI abdomen, MRI pelvis, and MRI neck. This resulted in a total of 58,405 reports for training, 7,413 reports for validation, and 13,057 reports for testing. Note that chest X-ray reports are excluded from this dataset to avoid data contamination with the MIMIC-CXR dataset for the report generation task. For each report, we used the same preprocessing functions as in~\cite{delbrouck2022vilmedic, delbrouck2022toward} to extract the findings and impression sections. Specifically, we filtered out the reports whose findings section are longer than 600 tokens. We performed a report summarization task by predicting the impression section given the findings section as input, which is another language-only task that does not require multi-modal input.

\subsection{Multimodal datasets}

\paragraph{PAD-UFES-20} consists of 2,298 clinical images of skin lesions collected from different smartphone devices with varying resolutions, sizes, and lighting conditions~\cite{pacheco2020pad}. The data was collected through the Dermatological and Surgical Assistance Program at the Federal University of Espírito Santo (UFES-Brazil), a nonprofit program that provides free skin lesion treatment. The dataset contains six different types of skin lesions including: Basal Cell Carcinoma (BCC), Malignant Melanoma (MEL), Squamous Cell Carcinoma (SCC), Actinic Keratosis (ACK), Melanocytic Nevus (NEV), and Seborrheic Keratosis (SEK). Each image is associated with up to 21 patient clinical features such as patient demographics, family cancer history lesion location, lesion size. We set up a 6-class classification task in a generative framework through a language decoder using skin lesion images and the associated clinical textual features as the multimodal input. Specifically, we selected 14 clinical attributes in the metadata for each lesion including: \textit{age}, \textit{gender}, \textit{smoke}, \textit{drink}, \textit{skin cancer history}, \textit{cancer history}, \textit{region}, \textit{fitspatrick}, \textit{horizontal and vertical diameters}, \textit{itch}, \textit{grew}, \textit{bleed}, and \textit{elevation}. The class ratio is approximately 16:1:4:14:5:4 over three skin cancers (BCC, MEL, and SCC) and three skin disease (ACK, NEV, and SEK). Since there are no published official train/test splits, we randomly split the dataset into a training set (80\%) and a test test (20\%) using a stratified sampling to the preserve original class ratio. We applied a series of image augmentation operations using RandAugment~\cite{cubuk2020randaugment} to the training set including: \textit{autoContrast}, \textit{equalize}, \textit{invert}, \textit{rotate}, \textit{posterize}, \textit{solarize}, \textit{color}, and \textit{contrast}.

\paragraph{VinDr-Mammo} is a full-field digital mammography dataset which consists of 5000 breast X-ray imaging studies and a total of 20,000 gray-scale images with extensive breast-level assessment and lesion-level annotations, collected from two hospitals in in Hanoi, Vietnam~\cite{nguyen2023vindr}. Each study contains four images where the left and right breasts are imaged with mediolateral-oblique (MLO) and cranio-caudal (CC) views. Each image has breast-level assessment following the Breast Imaging Reporting and Data System (BI-RADS). BI-RADS assessment ranges from 1 (negative) to 5 (highly suggestive of malignancy). In addition to the BI-RADS score, the breast density level is also provided as well as regional abnormality finding annotations. We performed a breast-level 5-class BI-RADS classification task similar to the setup in~\cite{wantlin2023benchmd}, except that the laterality and view position of the image was provided as additional contextual features. We used the official train/test splits where the train split contains 16,000 samples with a class ratio of 60:21:4:3:1 across BI-RADS 1-5, respectively and the test split contains 4,000 samples with the same class ratio. We applied the following transformations to the images in the training set: \textit{contrast}, \textit{equalize}, \textit{rotate}, \textit{shearX}, \textit{shearY}, \textit{translateX}, and \textit{translateY}. To mitigate the class imbalance in the training data, we upsampled for each minority class (BI-RADS 2-5) by a factor of 3. 

\paragraph{CBIS-DDSM} is the Curated Breast Imaging Subset of Digital Database for Screening Mammography~\cite{lee2017curated}. This dataset contains 2,620 scanned film mammography studies. Unlike VinDr-Mammo, CBIS-DDSM does not have breast-level BI-RADS assessment. Annotations are provided at the lesion level including BI-RADS, subtlety level, and pathology type. There are two types of lesions: mass and calcification. Both of them are annotated with three possible pathology labels: benign, benign without callback, and malignant. We performed a 3-class abnormality (patch-level) pathology classification task on this dataset for mass and calcification abnormalities separately. Abnormality image patch is cropped by the bounding box of the region-of-interest (ROI) from the full mammogram and used as the model input along with its view position (CC or MLO) information. We used the official train/test splits for both abnormality types. For mass cases, the training and test sets contain 1,318 and 378 images (class ratio: 6:1:6), respectively. For calcification cases, the total number of images in the training and test sets are 1,544 and 326 (class ratio: 1:1:1), respectively. For both cases, we applied the same image augmentation as in VinDr-Mammo to the training set.

\paragraph{PrecisionFDA Truth Challenge V2} was developed for benchmarking the state-of-the-art of variant calling in challenging genomics regions~\cite{olson2022pfda}. Genomic variant calling is a task aiming at identifying genetic variants from sequencing data~\cite{depristo2011gatk}, which can identify disease-causing mutations~\cite{aldubayan2020clincohort}. For variant calling, sequencing data is mapped to the coordinates of a reference genome~\cite{liao2023completeref}. The mappings can be represented as an image-like format that computational methods such as DeepVariant~\cite{poplin2018deepvariant} use to call variants, or in a human-friendly image format which experts use to inspect and quality control variants of interest~\cite{thorvaldsdottir2012igv}. For this task, we used an extensively characterized groundtruth set from the National Institute of Standards and Technology (NIST)~\cite{zook2016giab} for the HG002 sample. We generated examples from sequencing from the PrecisionFDA Truth Challenge V2. For training, we use 4\% of the examples from the whole genome (except for chromosome 20, 21, and 22). For evaluation, we used chromosome20, bases 3000001-9444417. This generated 197,038 candidate variants for training and 13,030 candidate variants for evaluation. For each example, the model predicts three possible genotypes, corresponding to how many copies (0, 1, or 2) of the given alternate allele are present. The training set consists of  45,011, 93,246, and 58,781 samples for classes 0, 1, 2, respectively. The evaluation set contains 3,016, 6,169, and 3,845 for classes 0, 1, 2, respectively.

We used DeepVariant v1.3.0's~\cite{poplin2018deepvariant} example generation method to create image-like examples suitable for machine classification. Specifically, input examples to DeepVariant v1.3.0 have a shape of (100, 221, 6) corresponding to (height, width, channels). Channels are shown in grey-scale below in the following order:
\begin{enumerate}
  \item Read base: different intensities represent A, C, G, and T.
  \item Base quality: set by the sequencing machine. White is higher quality.
  \item Mapping quality: set by the aligner. White is higher quality.
  \item Strand of alignment: Black is forward; white is reverse.
  \item Read supports variant: White means the read supports the given alternate allele, grey means it does not.
  \item Base differs from ref: White means the base is different from the reference, dark grey means the base matches the reference.
\end{enumerate}
To reshape the input example to be compatible with the Med-PaLM M input shape of (224, 224, 3), we stacked up channels 1, 2, 3 with channels 4, 5, 6 such that the original tensor of shape (100, 221, 6) became an RGB image of shape (200, 221, 3). We then padded the image on the width and height dimensions to give it a final shape of (224, 224, 3).

\paragraph{VQA-RAD} is a radiology visual question answering (VQA) dataset which consists of 315 radiology images and 3,515 question–answer pairs created and validated by clinicians~\cite{lau2018dataset}. The radiology images are selected from three imaging modalities (CT, MRI, and X-rays) and three anatomical regions (head, abdominal, chest). The types of question fall into 11 categories including modality, plane, organ system, abnormality, size, plane, positional reasoning, color, counting, attribute and other. 58\% of  the question–answer (QA) pairs are closed-ended (yes/no or limited choices) and the rest 42\% are open-ended (short answer). We adopted the official train/test splits, where the training set contains 1,797 QA pairs (only free-form and paraphrased questions were included) and the test set contains 451 QA pairs (not filtered).

\paragraph{Path-VQA} is a pathology VQA dataset, containing a total of 4,998 pathology images with 32,799 question-answer pairs~\cite{he2020pathvqa}. Pathology images are extracted from medical textbooks and online digital libraries. Each image is associated with multiple QA pairs pertaining to different aspects of the pathology including color, location, appearance, shape, etc. Open-ended questions account for 50.2\% of all questions, which are categorized into 7 categories: what, where, when, whose, how, and how much/how many, accounting for 50.2\% of all questions. The rest are close-ended questions with simple "yes/no" answer. We adopted the official data partitioning where the training, validation, and test sets contain 19,755, 6,279, and 6,761 QA pairs, respectively.

\paragraph{Slake-VQA} is a semantically annotated and knowledge-enhanced bilingual (English and Chinese) VQA dataset on radiology images~\cite{liu2021slake}. It contains 642 annotated images with 14,028 question-answer pairs covering 12 diseases, 39 organ systems and 3 imaging modalities (CT, MRI, and chest X-rays). Questions are either open-ended (free-form) or closed-ended (balanced yes/no) related to various aspects of the image content including plane, quality, position, organ, abnormality, size, color, shape, knowledge graph, etc. The training, validation, and test sets contain 9,849, 2,109, and 2,070 samples, respectively.

\paragraph{MIMIC-CXR} is a large dataset of chest radiographs with free-text radiology reports~\cite{johnson2019mimic}. A total of 377,110 images are available in the dataset from 227,835 image studies collected for 65,379 patients. Each patient may have multiple studies and each study may contain one or more images associated with the same free-text report. Images in MIMIC-CXR are collected from multiple view positions: e.g., anterior-posterior (AP), posterior-anterior, and lateral (LA). Protected health information (PHI) in radiology reports and images is removed, which results in missing information in some sentences of the reports. Since this dataset contains sequential imaging studies of an individual patient, a large number of reports refer to information in prior studies of the same patient. Each report is annotated with structured labels of 14 common radiological observations using CheXpert labeler~\cite{irvin2019chexpert}. We performed two tasks using this dataset: chest X-ray report generation and binary classification of clinically-relevant pathology observations. We preprocessed the radiology reports by extracting the indication, findings, and impression sections, removing redundant white-spaces in the reports, following previous work~\cite{chen2020generating}. We used the official train/validation/test splits. We discarded images without reports and reports where the findings section can not be extracted across train and test. We also filtered out the reports where the length of findings section exceeds 800 characters. However, unlike most previous work using focusing only on the frontal view, we treated images of different orientation that are associated with the same report as independent samples (retaining the patient-level train/test splits to avoid contamination of the test data). The goal is to improve the image understanding capability of the model to process images of different view positions. In a separate evaluation, we also studied a subset of samples where reports are accompanied by both a front and lateral view (two-view report generation).

For the report generation task, we combined the chest X-ray image with the contextual information from the indication section (reason for the study) to predict the findings section of the target report. The total number of samples in the training, validation, and test sets are: 353,542, 2,866, and 4,834, respectively.

For the binary classification task, we grouped negative and uncertain labels as the negative class for 11 pathological conditions: no finding, atelectasis, cardiomegaly, consolidation, edema, pleural effusion, lung opacity, enlarged cardiomediastinum, fracture, pneumonia, and support devices. Atelectasis, cardiomegaly, consolidation, edema, and pleural effusion are 5 major conditions given their clinical relevance and prevalence. The "No finding" label captures the cases without any pathology and therefore this classification task simply helps the model to distinguish normal cases from cases with any type of abnormality. Due to class imbalance, during training we upsampled the positive class by a factor of 2 for the following conditions: consolidation, enlarged cardiomediastinum, fracture, and pneumonia. These binary classification tasks are auxiliary to the report generation task when they are trained simultaneously since they help the model to distinguish among different types of clinical observations in the chest X-ray images.

\begin{figure}[t]
\centering
     \includegraphics[width=0.8\textwidth]{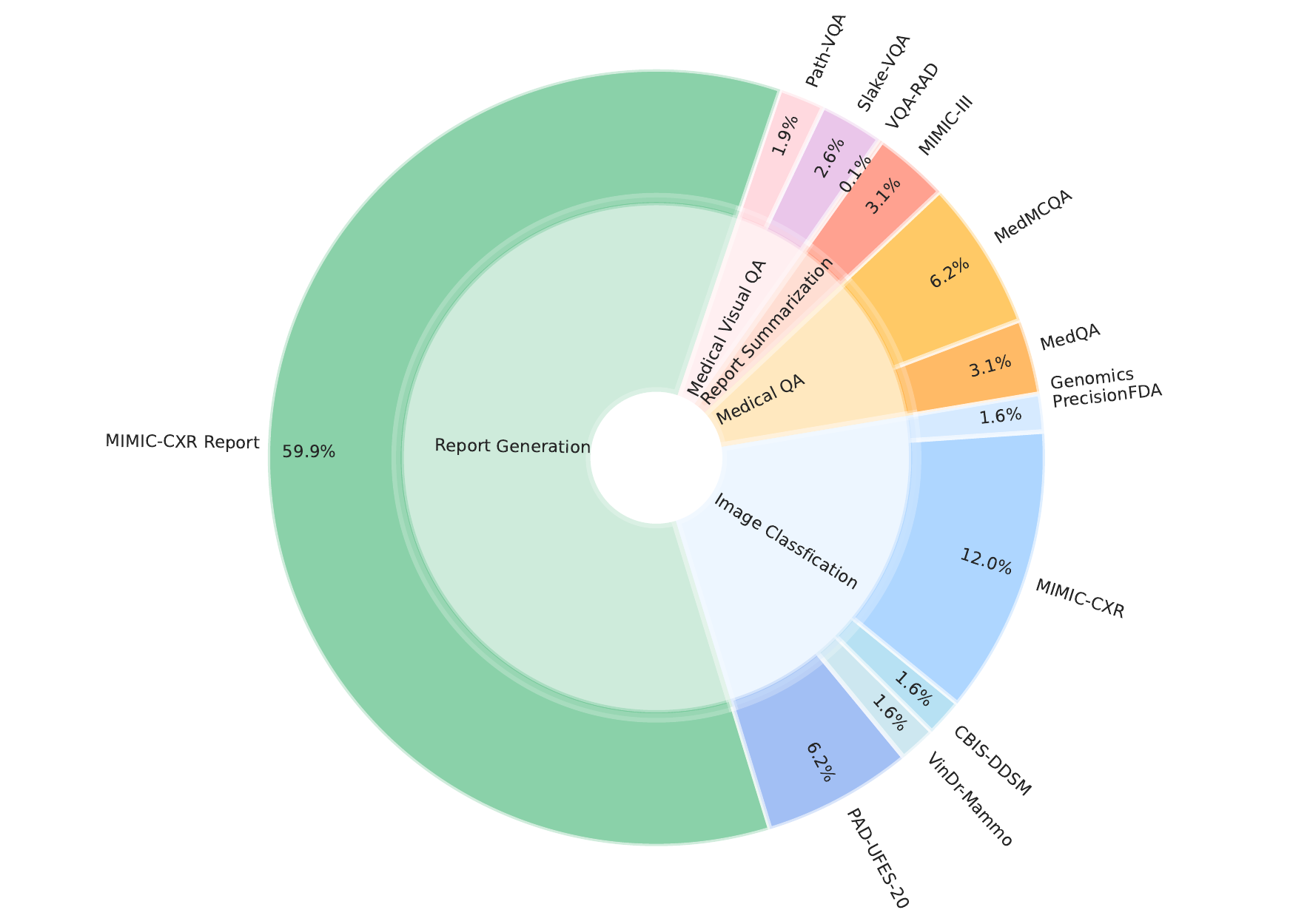}
     \vspace{6pt}
     \caption{\textbf{Med-PaLM M data mixture overview.} Summary of the mixture ratio in the Med-PaLM M training data mixture across MultiMedBench datasets as detailed in Table~\ref{tab-appendix:data mixture}. }
     \label{fig:datasets-mixture-overiew}
\end{figure}

\section{Med-PaLM M Training Details}
\subsection{Training data mixture}
\cref{fig:datasets-mixture-overiew,tab-appendix:data mixture} show the mixture ratio and few-shot task setup of the training data mixture. The majority of the
data distribution is medical vision-language tasks, with less than 15\% consisting of language-only tasks. While the majority of vision-language tasks were trained with a text-only 1-shot setup (without the corresponding image), the CBIS-DDSM classification and genomic variant calling tasks were trained with a 0-shot setup.
\begin{table}[ht]
\small
\centering
\caption{\textbf{Med-PaLM M data mixture.} Summary of the task types, modalities, mixture ratios, and few-shot setups in Med-PaLM M training data mixture.}
\label{tab-appendix:data mixture}
\begin{tabular}{ccccc}
\toprule
Task & Modality & Dataset          & Mixture ratio  & Few-shot setup  \\ \midrule
\multirow{2}{*}{Question Answering} & \multirow{2}{*}{Text}
& MedQA       & 3.13\%       & 2-shot  \\ 
& & MedMCQA     & 6.25\%        & 2-shot \\ 
\midrule
Report Summarization & Radiology & MIMIC-III       & 3.13\% & 0-shot   \\ 
\midrule
\multirow{3}{*}{\begin{tabular}[c]{@{}c@{}}Visual\\Question Answering\end{tabular}} & \multirow{2}{*}{Radiology}
& VQA-RAD & 0.15\%  & text-only 1-shot  \\
&& Slake-VQA & 2.64\% & text-only 1-shot \\
& Pathology & Path-VQA  & 1.90\%  & text-only 1-shot  \\
\midrule
Report Generation &Chest X-ray &MIMIC-CXR         & 59.90\% & text-only 1-shot  \\
\midrule
\multirow{6}{*}{\begin{tabular}[c]{@{}c@{}}Medical\\Image Classification\end{tabular}}
& Dermatology &PAD-UFES-20  & 6.25\%   & text-only 1-shot  \\
& \multirow{2}{*}{Mammography}
& VinDr-Mammo         & 1.56\% & text-only 1-shot  \\
&& CBIS-DDSM          & 1.56\% & 0-shot  \\
& Chest X-ray & MIMIC-CXR        & 11.98\%  & text-only 1-shot  \\ 
& {\begin{tabular}[c]{@{}c@{}}Genomics\end{tabular}} & {\begin{tabular}[c]{@{}c@{}}PrecisionFDA\\Truth Challenge V2~\cite{olson2022pfda}\end{tabular}}     & 1.56\% & 0-shot  \\

\bottomrule 
\end{tabular}
\end{table}

\subsection{Training hyperparameters}
PaLM-E projects the multimodal inputs into the same language embedding space as latent vectors such that continuous observations (e.g., images, time series) can be processed the same way by a pre-trained LLM as the language tokens, and thereby generates textual completions autoregressively given a multimodal prompt. In our experiments, the ViT maps the visual input to a fixed number of 256 tokens which are further processed by the LLM along with the additional text/multimodal tokens~\cite{driess2023palme}. 
Med-PaLM M was finetuned on the pretrained PaLM-E checkpoints. \cref{tab-appendix:finetuning hyperparameters} shows the training hyperparameters for Med-PaLM M 12B, 84B, and 562B, respectively. 

\begin{table}[]
\small
\centering
\caption{\textbf{Med-PaLM M finetuning hyperparameters.} Summary of the finetuning hyperparameters for Med-PaLM M 12B, 84B, and 562B.}
\label{tab-appendix:finetuning hyperparameters}
\begin{tabular}{cccc}
\toprule
Hyperparameter            & Med-PaLM M (12B) & Med-PaLM M (84B) & Med-PaLM M (562B)  \\ \midrule
Learning rate & $5\times 10^{-5}$ & $5\times 10^{-5}$ & $2.5\times 10^{-5}$   \\
Batch size & 128 & 128 & 256   \\
Max token input length & 710 & 710 & 710\\
Max token output length & 256 & 256 & 256\\

\bottomrule 
\end{tabular}
\end{table}
\section{Detailed Med-PaLM M Performance}
\label{appendix:multimedbench_detailed_performance}

\paragraph{Performance on text-only medical question answering}
We report the few-shot performance of Med-PaLM M on MedQA, MedMCQA, and PubMedQA in Table~\ref{tab:results-med-qa}. SOTA results were chosen from Med-PaLM 2 with ensemble refinement prompting and PaLM 540B few-shot results reported in~\cite{singhal2022large, singhal2023towards}. Med-PaLM M outperformed the baseline PaLM model (from which it inherits) by a large margin on all three datasets, despite falling behind the Med-PaLM 2 best results obtained with ensemble refinement. Scaling up the language model from 8B to 540B significantly improves the accuracy on the multiple-choice medical question answering tasks, where strong capabilities to comprehend, recall, and reason about medical knowledge are important. These results can be partly explained by the improved base language model used for Med-PaLM 2.

\begin{table}[ht]
\small
\centering
\caption{\textbf{Language-only medical question answering accuracy on MultiMedQA.} Med-PaLM 2 results with ensemble refinement \cite{singhal2023towards} and PaLM few-shot results \cite{singhal2022large} are presented for comparison. Few-shot Med-PaLM~M outperforms the corresponding PaLM baseline by a large margin, despite falling short of the state-of-the-art Med-PaLM 2.}
\label{tab:results-med-qa}
\begin{tabular}{@{\hspace{.01cm}}ccccccc@{\hspace{.01cm}}}
\toprule
Dataset       & Med-PaLM 2   & PaLM   & Med-PaLM M (12B) & Med-PaLM M (84B) & Med-PaLM M (562B) \\ \midrule
MedQA (USMLE) & \textbf{86.50\%} & 58.90\% & 29.22\%        & 46.11\%         & 69.68\% \\
\midrule
MedMCQA       & \textbf{72.30\%} & 54.50\% & 32.20\%         & 47.60\%          & 62.59\% \\
\midrule
PubMedQA      & \textbf{81.80\%} & 55.00\%  & 48.60\%         & 71.40\%          & 80.00\%    \\ \bottomrule
\end{tabular}
\end{table}

\paragraph{Performance on radiology report summarization}
We report commonly used metrics such as ROUGE-L~\cite{lin2004rouge}, BLEU~\cite{papineni2002bleu}, and F1-RadGraph~\cite{jain2021radgraph} scores on the radiology report summarization task as in~\citet{van2023radadapt} in Table~\ref{tab:results-report-summarization}. Med-PaLM M (562B) yielded the best overall performance compared to the smaller model variants, consistent with our observations on medical question answering tasks. Med-PaLM M performed worse than the SOTA results which were obtained with a parameter-efficient finetuning method (low-rank adaptation, LoRA~\cite{hu2021lora}) on a 738M-parameter clinical-T5 model~\cite{lehman2023clinical}. However, as noted in~\cite{van2023radadapt}, one caveat of clinical-T5 is that it is unclear if~\citet{lehman2023clinical} pretrained the model on the test set of MIMIC-III which led to potential data leakage. Notably, Med-PaLM M compared favorably to the results in~\citet{van2023radadapt} based on the T5 model which was not pretrained on clinical text, similar to the PaLM model.

\begin{table}[ht]
\small
\centering
\caption{\textbf{Med-PaLM M performance on MIMIC-III radiology report summarization.}} 
\label{tab:results-report-summarization}
\begin{tabular}{cccccc}
\toprule
Dataset                    & Metric      & SOTA     & Med-PaLM M (12B) & Med-PaLM M (84B) & Med-PaLM M (562B) \\ \midrule
\multirow{3}{*}{MIMIC-III} & ROUGE-L     & \textbf{38.70}\%  & 29.45\%        & 31.47\%         & 32.03\% \\
& BLEU        & \textbf{16.20}\%   & 12.14\%        & 15.36\%         & 15.21\% \\
                           & F1-RadGraph & \textbf{40.80}\%   & 31.43\%        & 33.96\%         & 34.71\% \\
 \bottomrule
\end{tabular}
\end{table}

\paragraph{Performance on medical image classification tasks}
Table~\ref{tab:results-classifcation} shows the performance of Med-PaLM M on a set of classification tasks across multiple modalities including dermatology, radiology, and genomics. Since these tasks all have imbalanced class distributions, we reported macro-AUC (unweighted mean of all the per-class AUC scores) and macro-F1 scores (unweighted mean of all the per-class F1 scores) as the classification metrics except for the genomic variant calling task where the F1 scores for single nucleotide polymorphisms (SNPs) and short insertions and deletions (indels) in the context of variant discovery were used instead. 

On VinDr-Mammo, all size variants of Med-PaLM M exceeded the prior SOTA using a smaller ViT (9.7M) on macro-AUC~\cite{wantlin2023benchmd}. On CBIS-DDSM, our model achieved the best macro-F1 of 51.12\% and 67.86\% on the mass and calcification classification, respectively, behind the SOTA F1 of 70.71\% reported on the calcification classification~\cite{panambur2022effect}. Note that most previous works on CBIS-DDSM focused on a two-class patch-level classification (benign versus malignant) problem in contrast to our 3-class setup as discussed in~\cite{petrini2022breast}. On Pad-UFES-20, since no official train/test splits are available, our results are not directly comparable with prior studies. Med-PaLM M 84B achieved a macro-AUC of 97.27\%, on par with previous reported results (94\% - 98\%) obtained using CNN and ViT variants~\cite{dai2022deeply, de2022exploring}. On MIMIC-CXR, we reported the macro-average of F1 scores across the binary classification of 5 major conditions: atelectasis, cardiomegaly, consolidation, edema, and pleural effusion. Med-PaLM M (562B) achieved a macro-AUC of 79.09\%, slightly lower than the SOTA result of 81.27\% obtained from ParallelXNet~\cite{rammuni2022effective}, which used a parallelization of various CNN Architectures. On the variant calling task, DeepVariant model~\cite{poplin2018deepvariant} outperformed Med-PaLM M on both Indel-F1 and SNP-F1 scores. The SOTA DeepVariant model was trained with 2,633-fold more training examples. Training with the same examples resulted in a narrower advantage for DeepVariant for SNP (Med-PaLM M 99.35\% versus DeepVariant 99.63\%) and Indel (Med-PaLM M 97.04\% versus DeepVariant 98.55\%. Notably, Med-PaLM M outperformed the accuracy of the widely used GATK4 method~\cite{depristo2011gatk} for SNP calling (Med-PaLM M 99.35\% versus GATK4 99.29\%) but not Indel calling (Med-PaLM M  97.04\% versus GATK4 99.32\%).

Taken together, Med-PaLM M achieved competitive results on a variety of classification tasks using a single model compared to highly specialized SOTA models. It is worth noting that we did not perform any fine-grained task-specific customization and hyperparameter tuning beyond data augmentation and class balancing. It is expected that scaling up the language model does not significantly benefit the classification tasks where the vision encoder is likely the bottleneck for the model performance. There is no overall evidence to suggest that larger vision model outperforms the small one across all our experiments, suggesting that more domain-specific pretraining may be more important for improving vision encoder performance. It is also likely that relatively small-scale datasets we explored here are not sufficient to establish such a robust scaling relationship between the model size and task performance, as results were generally close to each other across model scales.

\begin{table}[ht]
\small
\centering
\caption{\textbf{Med-PaLM M performance on medical image classification.} We report macro-averaged AUC and F1 for all tasks. For MIMIC-CXR, metrics are averaged over 5 major pathological conditions.}
\label{tab:results-classifcation}
\begin{tabular}{ccccccc} 
\toprule
Dataset &
  \# Classes &
  Metric &
  SOTA &
  \begin{tabular}[c]{@{}c@{}}Med-PaLM M \\ (12B)\end{tabular} &
  \begin{tabular}[c]{@{}c@{}}Med-PaLM M \\ (84B)\end{tabular} &
  \begin{tabular}[c]{@{}c@{}}Med-PaLM M \\ (562B)\end{tabular} \\ \midrule
\multirow{2}{*}{\begin{tabular}[c]{@{}c@{}}MIMIC-CXR\\ (5 conditions)\end{tabular}}     & \multirow{2}{*}{2-class} & Macro-AUC & \textbf{81.27\%}    & 76.67\%          & 78.35\%          & 79.09\% \\
                                 &                           & Macro-F1  & N/A       & 38.33\%          & 36.83\%          & \textbf{41.57\%} \\ \midrule
\multirow{2}{*}{PAD-UFES-20}   & \multirow{2}{*}{6-class}  & Macro-AUC & N/A & 95.57\%          & \textbf{97.27\%} & 96.08\%          \\
                                 &                           & Macro-F1  & N/A       & 78.42\%          & \textbf{84.32\%} & 77.03\%          \\ \midrule
\multirow{2}{*}{Variant Calling} & \multirow{2}{*}{3-class}  & Indel-F1  & \textbf{99.40}\%   & 96.42\%          & 97.04\% & 95.46\%          \\
                                 &                           & SNP-F1    & \textbf{99.70}\%   & 99.35\% & 99.32\%          & 99.16\%          \\ \midrule
\multirow{2}{*}{VinDr-Mammo}    & \multirow{2}{*}{5-class}  & Macro-AUC & 64.50\%    & 66.29\%          & \textbf{71.76\%} & 71.42\%          \\
                                 &                           & Macro-F1  & N/A       & 29.81\%          & \textbf{35.7\%}           & 33.90\%           \\ \midrule
\multirow{2}{*}{\begin{tabular}[c]{@{}c@{}}CBIS-DDSM\\ (mass)\end{tabular}} &
  \multirow{2}{*}{3-class} &
  Macro-AUC &
  N/A &
  70.11\% &
  73.09\% &
  \textbf{73.31\%} \\
                                                   
                                 &                           & Macro-F1  & N/A       & 47.23\%          & 49.98\%          & \textbf{51.12\%} \\ \midrule
\multirow{2}{*}{\begin{tabular}[c]{@{}c@{}}CBIS-DDSM\\ (calcification)\end{tabular}} &
  \multirow{2}{*}{3-class} &
  Macro-AUC &
  N/A &
  81.40\% &
  \textbf{82.22\%} &
  80.90\% \\
                                 &                           & Macro-F1  & \textbf{70.71\%}       & 67.86\%          & 63.81\% & 63.03\%          \\ \bottomrule
\end{tabular}
\end{table}

\paragraph{Performance on medical visual question answering} 
Since we formulated both close-end and open-end QA pairs in three VQA datasets as an open-ended language decoding task conditioned on visual input, we used BLEU-1 and token-level F1 scores to assess the performance of Med-PaLM M. This is in contrast with many prior works which used a string-level accuracy evaluation metric as they often considered VQA as a classification task on a set of pre-defined fixed-number answer candidates~\cite{liu2023q2atransformer, eslami2021does}. This accuracy metric has the weakness of failing to capture "near misses" of groundtruth answers, particularly in our open-ended generative setup. We also noted that only human validation by experts can provide additional insights on the quality of model answers beyond token-level or string-level matching metrics. As shown in Table~\ref{tab:results-vqa}, Med-PaLM M surpassed previous SOTA using a similar generative approach across all three datasets and metrics~\cite{van2023open, bazi2023vision}. In particular, model performance increased with scaling up the language model on VQA-RAD and Path-VQA. On Slake-VQA, the best performance was achieved with the medium size model variant. These results suggest that scaling up language models is beneficial for visual-language tasks where language reasoning is conditioned on visual understanding.

\begin{table}[ht]
\small
\centering
\caption{\textbf{Med-PaLM M performance on medical visual question answering.} Med-PaLM exceeds prior SOTA on all three VQA tasks.}
\label{tab:results-vqa}
\begin{tabular}{cccccc}
\toprule
Dataset                    & Metric      & SOTA    & Med-PaLM M (12B)    & Med-PaLM M (84B)  & Med-PaLM M (562B) \\ \midrule
\multirow{2}{*}{VQA-RAD}   & BLEU-1    & 71.03\%  & 64.02\%        & 69.38\%          & \textbf{71.27\%} \\
                          
                           & F1          & N/A     & 50.66\%        & 59.90\%           & \textbf{62.06\%} \\ \midrule
\multirow{2}{*}{Path-VQA}  & BLEU-1      & 70.30\%  & 68.97\%        & 70.16\%          & \textbf{72.27\%} \\
                           
                           & F1          & 58.40\%  & 57.24\%        & 59.51\%          & \textbf{62.69\%} \\ \midrule
\multirow{2}{*}{Slake-VQA} & BLEU-1      & 78.60\%  & 90.77\%        & \textbf{92.7\%}  & 91.64\%          \\
                           & F1          & 78.10\%  & 86.22\%        & \textbf{89.28\%} & 87.50\%           \\ \bottomrule
\end{tabular}
\end{table}

\paragraph{Performance on chest X-ray report generation} 
To measure the quality of generated chest X-ray reports using automatic metrics, we computed common natural language generation metrics such as BLEU-1, BLEU-4, ROUGE-L, CIDEr-D~\cite{vedantam2015cider}, in addition to the clinical efficacy (CE) metrics and F1-RadGraph which were designed to capture the factuality and diagnostic accuracy in the generated reports. Specifically, we used CheXbert~\cite{smit2020chexbert}, an automatic radiology report labeller based on a BERT model improved with expert annotation, to extract the 14 CheXpert pathological observations from a given report. For each observation, the predicted label was compared against the groundtruth label to compute CE metrics. F1-RadGraph generalizes over CheXbert labeller to more observation categories by measuring the overlapping clinical entities and relations between a generated report and the reference report~\cite{yu2022evaluating}. In line with previous studies~\cite{liu2019clinically, chen2020generating, miura2020improving, nicolson2022improving, bannur2023learning, tanida2023interactive, jeong2023multimodal}, we reported the macro-F1 and micro-F1 scores averaged over 5 major observations and all 14 observations for CE metrics, respectively. As shown in Table~\ref{tab:results-cxr-report-gen}, Med-PaLM M achieved a new SOTA on all CE metrics and F1-RadGraph, with a substantial increase of about 9 points on macro-F1-14 and micro-F1-14 averaged across all clinical relevant observations over previous best SOTA results in~\cite{nicolson2022improving, jeong2023multimodal}. The macro-average F1 resulted in a lower score than the micro-average F1 over 14 observation categories because of the worse model performance on some categories with very low representation in the training data. Notably, improvements on F1 scores were more prominent across all 14 categories than over the 5 major categories for Med-PaLM M. This is likely due to the benefit of jointly training with the classification tasks on those minority conditions. We consider such positive task transfer as one of the main advantages of a generalist multi-task model over a specialized single-task model. On text overlap based natural language generation metrics, Med-PaLM M did not outperform existing SOTA results. However, the pitfalls of these automatic metrics have been raised by many studies, particularly in that they fail to capture the factual correctness and do not align well with radiologist judgements~\cite{liu2019clinically, miura2020improving, yu2022evaluating, tanida2023interactive}. 

Interestingly, our largest model Med-PaLM M (562B) did not achieve the best performance, falling slightly short of Med-PaLM M (84B). Furthermore, the gap in performance across three model sizes is relatively small across all types of metrics. The diminishing return of increasing the size of the language model is likely because the output space for chest X-ray report generation is fairly confined to a set of template sentences and limited number of conditions. It is also possible that the task performance is primarily limited by the vision encoder, particularly in how well it is adapted for this domain. As noted by~\citet{xu2021vitae}, ViT lacks inductive bias for modeling local visual features which are often crucial for interpreting medical images. To overcome this limitation, large-scale medical training data may be required to enable benefit from size scaling. Additionally, the input image size $224 \times 224 \times 3$ we used cause loss in resolution, which is a tradeoff we made to shorten the length of embedded image tokens to fit within the context length limit of the language model.

\begin{table}[ht]
\small
\centering
\caption{\textbf{Med-PaLM M performance on chest X-ray report generation.} We use text-overlap based and clinical factuality based automatic metrics to evaluate the quality of model-generated reports. Med-PaLM M sets new SOTA on all metrics designed to capture clinical efficacy and correctness. Across three Med-PaLM M variants, the medium-sized model achieves the best performance.}
\label{tab:results-cxr-report-gen}
\begin{tabular}{ccccc}
\toprule
Metric              & SOTA    & Med-PaLM M (12B)    & Med-PaLM M (84B)  & Med-PaLM M (562B) \\ \midrule
Micro-F1-14 & 44.20\%  & 51.41\%        & \textbf{53.56\%} & 51.60\% \\
Macro-F1-14 & 30.70\%  & 37.31\%        & \textbf{39.83\%} & 37.81\% \\
Micro-F1-5  & 56.70\%  & 56.54\%        & \textbf{57.88\%} & 56.28\% \\
Macro-F1-5  & N/A     & 50.57\%        & \textbf{51.60\%} & 49.86\% \\
F1-RadGraph         & 24.40\%  & 25.20\%        & \textbf{26.71\%} & 26.06\%          \\
BLEU-1           & \textbf{39.48\%} & 30.90\%        & 32.31\% & 31.73\% \\
BLEU-4           & \textbf{13.30\%} & 10.43\%        & 11.31\%          & 11.50\%  \\
ROUGE-L             & \textbf{29.60\%} & 26.16\%        & 27.29\% & 27.49\% \\
CIDEr-D               & \textbf{49.50\%} & 23.43\%        & 26.17\% & 25.27\%          \\
 \bottomrule 
\end{tabular}
\end{table}

\section{Details on Human Evaluation}

\begin{figure}[ht]
    \centering
    \includegraphics[width=0.9\textwidth]{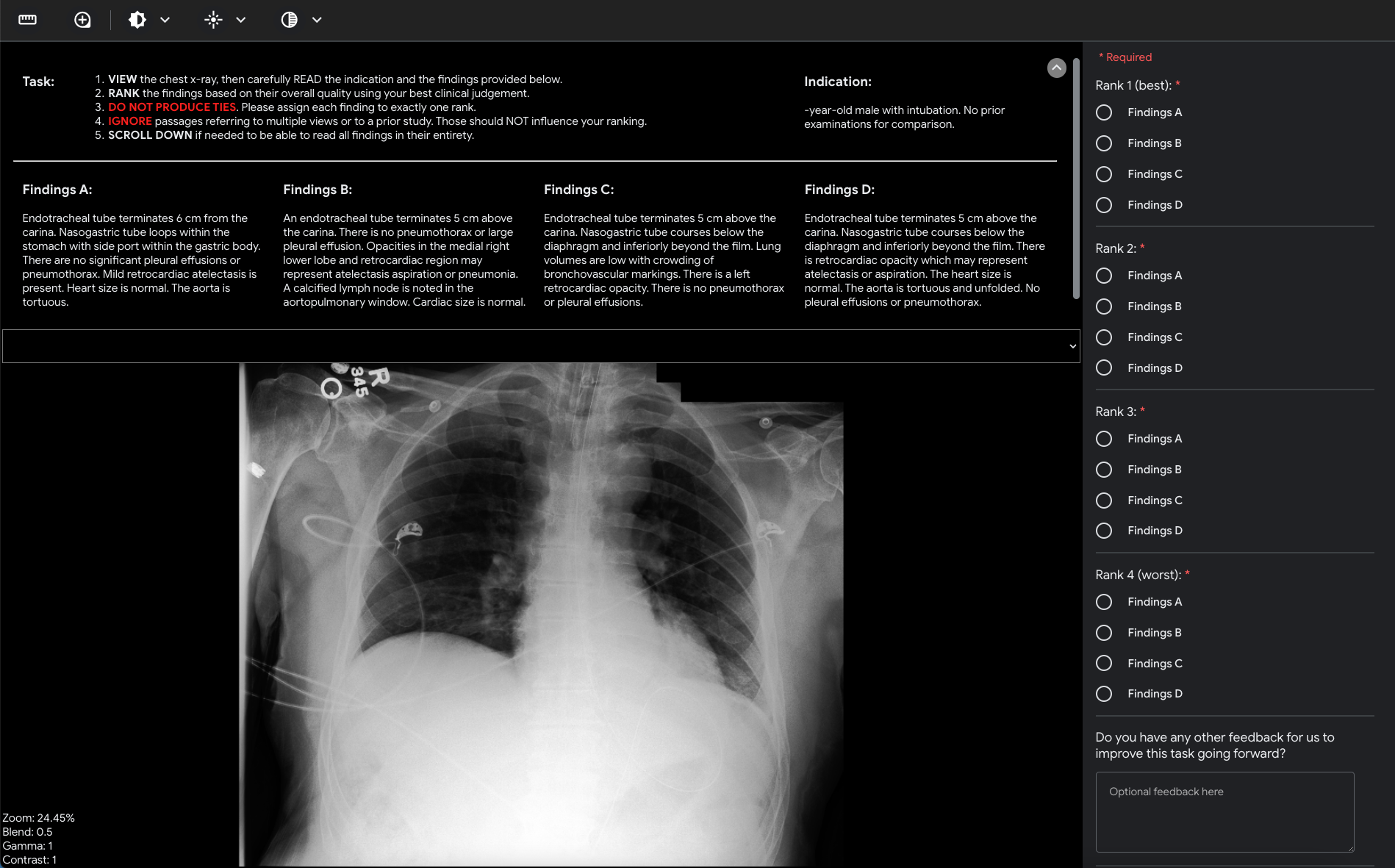}
    \vspace*{3mm}
    \caption{\textbf{Side-by-side human evaluation task interface.} Radiologist raters ranked four findings paragraphs based on overall quality, given a chest X-ray and indication. The four findings corresponded to the reference findings, and findings generated by three Med-PaLM M model variants (12B, 84B, 562B).}
    \label{fig-appendix:human-evaluation-ui-side-by-side}
\end{figure}

\begin{figure}[ht!]
    \centering
    \includegraphics[width=0.9\textwidth]{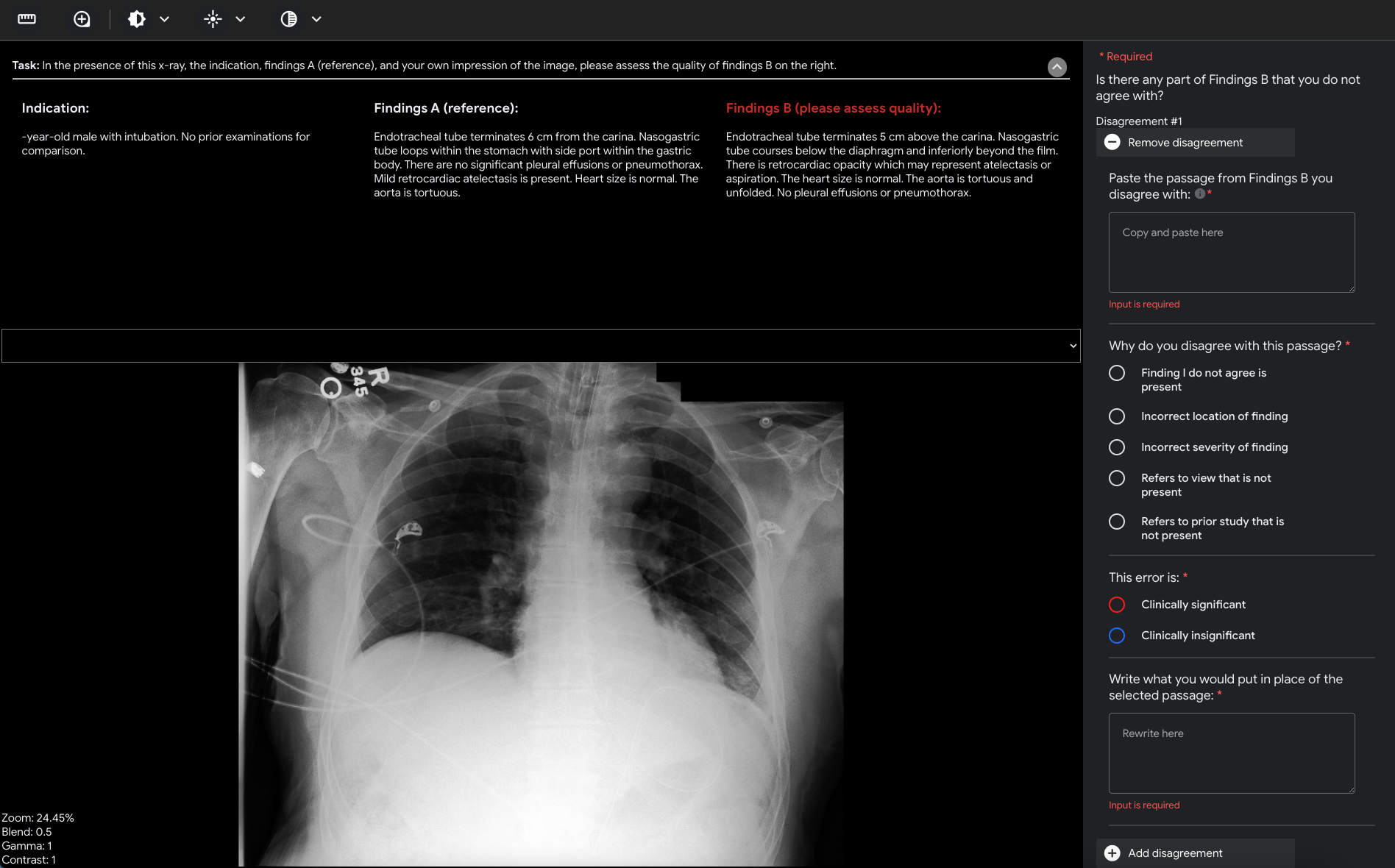}
    \vspace*{3mm}
    \caption{\textbf{Independent human evaluation task interface.} Radiologist raters annotated a findings paragraph generated by Med-PaLM M (red) for errors and omissions, given a chest X-ray, the indication, and reference findings.}
    \label{fig-appendix:human-evaluation-ui-independent}
\end{figure}

Figures \ref{fig-appendix:human-evaluation-ui-side-by-side} and \ref{fig-appendix:human-evaluation-ui-independent} depict the task interfaces used for side-by-side and independent radiologist evaluations, including the task input and annotation prompts presented to radiologist raters.
For ease of a detailed inspection (e.g., identification of subtle structures), the built-in medical image viewer provided tools for raters to adjust the chest X-ray image, including zoom, gamma, and blend controls.

It is worth noting that a non-trivial number of reports in the MIMIC-CXR dataset contain references to prior studies (e.g., ``Compared to the prior radiograph [...]'') or references to multiple views (e.g., ``Ap and lateral views of the chest are compared.''). By contrast, the input to our model is a single image and indication from a single study. As a result, these artifacts in the training corpus render the model prone to hallucination of references to non-existent prior imaging studies or non-existent X-ray views.
Our human evaluation task interface accounted for the presence of these artifacts by providing the option to categorize erroneous passages as ``Refers to view that is not present'' or ``Refers to study that is not present''.
Future work may leverage the CXR-PRO dataset~\cite{ramesh2022improving}, a cleaned version of MIMIC-CXR with all prior references removed, to mitigate this issue during model development. 

For the purpose of analysis, we distinguished between clinical errors (i.e., ``Finding I do not agree is present'', ``Incorrect location of finding'') and non-clinical errors (i.e., ``Refers to view that is not present'' or ``Refers to study that is not present'').
Table \ref{tab-appendix:human-evaluation-results} summarizes the rates of omissions and errors identified by clinician raters in radiology reports generated by different Med-PaLM M models. Here, we report the rate of total errors, including all clinical and non-clinical error types. On average, the best performing Med-PaLM M model produces 0.58 total errors per report.

One important limitation of our human evaluation approach is the inter-rater variability.
Similar to~\cite{yu2022evaluating}, which used a comparable evaluation scheme, we also observed that the same radiology report often was annotated with varying error and omission passages by different radiologist raters.
While this is a common phenomenon in studies that use subjective ratings from clinicians, future work may aim to further refine rater instructions and improve rater calibration to reduce variability.

\begin{table}[ht]
\small
\centering
\caption{\textbf{Independent human evaluation details.} Rates of omissions and errors identified by clinician raters in radiology reports generated by different Med-PaLM M models. Clinical errors are those related to the presence, location or severity of a clinical finding. Total errors include both clinical errors and non-clinical errors (i.e., passages referring to views or prior studies not present).}
\label{tab-appendix:human-evaluation-results}
\begin{tabular}{cccc}
\toprule
 Model Size                              & Med-PaLM M (562B)  & Med-PaLM M (84B)   & Med-PaLM M (12B)                 \\
\midrule
Significant Omissions                    & 0.10 (95\% CI, 0.08 - 0.12)  & 0.09 (95\% CI, 0.07 - 0.10)  & 0.08 (95\% CI, 0.06 - 0.10)  \\
Total Omissions                          & 0.13 (95\% CI, 0.11 - 0.16)  & 0.12 (95\% CI, 0.10 - 0.15)  & 0.12 (95\% CI, 0.10 - 0.15)  \\  \midrule
Significant Clinical Errors              & 0.26 (95\% CI, 0.23 - 0.29)  & 0.23 (95\% CI, 0.20 - 0.27)  & 0.26 (95\% CI, 0.22 - 0.29) \\
Total Clinical Errors                    & 0.29 (95\% CI, 0.25 - 0.32)  & 0.25 (95\% CI, 0.22 - 0.28)  & 0.28 (95\% CI, 0.24 - 0.31) \\
Total Errors                             & 0.63 (95\% CI, 0.58 - 0.68)  & 0.59 (95\% CI, 0.54 - 0.64)  & 0.58 (95\% CI, 0.53 - 0.63) \\
\bottomrule
\end{tabular}
\end{table}

\section{MultiMedBench Examples}

In \cref{tab:examples--multimedbench-classification,tab:examples--multimedbench-vqa-report-gen,tab:examples--report-summarization,tab:examples--medmcqa,tab:examples--medqa} we provide examples of various MultiMedBench tasks.

\begin{table}[ht]
\footnotesize
\caption{\textbf{Examples of the classification tasks in MultiMedBench.}}
\centering
\begin{tabular}{p{0.15\linewidth} p{0.6\linewidth}  p{0.1\linewidth}}
\toprule
Image & Task and input prompt & Target\\ \midrule

\raisebox{-1.8\height}{\includegraphics[scale=0.3]{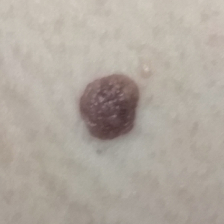}} 
& \textbf{Classification (PAD-UFES-20)}

\textbf{Instructions:} You are a helpful dermatology assistant. The following are questions about skin lesions. Categorize the skin lesions into the most likely class given the patient history.

Given <img>. Patient History: Age: 51, Gender: female, Smoke: false, Drink: false, Family skin cancer history: true, Family any cancer history: false, Lesion region: back, Lesion itch: false, Lesion grew: false, Lesion bled: false, Lesion elevation: false, Fitzpatrick scale: 1.0, Diameters (mm): [12.0, 8.0]. 
Q: Which of the following is the most likely diagnosis of the patient's skin lesion? 
(A) Nevus (B) Basal Cell Carcinoma (C) Squamous Cell Carcinoma (D) Actinic Keratosis (E) Seborrheic Keratosis (F) Melanoma 
A: Basal Cell Carcinoma.

Given \textbf{<img>}. Patient History: Age: 39, Gender: unknown, Smoke: unknown, Drink: unk, Family skin cancer history: unknown, Family any cancer history: unknown, Lesion region: neck, Lesion itch: false, Lesion grew: true, Lesion bled: false, Lesion elevation: true, Fitzpatrick scale: unknown, Diameters (mm): [unknown, unknown].
Q: Which of the following is the most likely diagnosis of the patient's skin lesion? 
(A) Nevus (B) Basal Cell Carcinoma (C) Squamous Cell Carcinoma (D) Actinic Keratosis (E) Seborrheic Keratosis (F) Melanoma
A:
& Nevus. \\

\midrule
\raisebox{-1.1\height}{\includegraphics[scale=0.3]{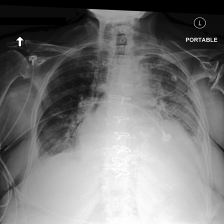}} 
& \textbf{Classification (MIMIC-CXR)}

\textbf{Instructions:} You are a helpful radiology assistant. The following are questions about findings in chest X-ray in different views. Identify if a specific type of abnormality is shown in the X-ray.

Given the AP view X-ray image <img>. Q: Is cardiomegaly indicated by the image?
(A) No (B) Yes

A: Yes.

Given the AP view X-ray image \textbf{<img>}. Q: Is cardiomegaly indicated by the image?
(A) No (B) Yes

A:
& Yes. \\
\midrule

\raisebox{-1.2\height}{\includegraphics[scale=0.3]{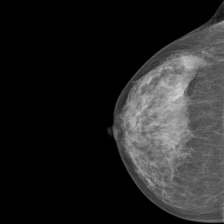}} 
& \textbf{Classification (VinDr-Mammo)} 

\textbf{Instructions:} You are a helpful medical assistant. The following are questions about mammography reading. Provide a breast-level assessment based on the BI-RADS categories.

Given mammogram image <img>. Image view: bilateral craniocaudal
Q: What is the most likely breast BI-RADS score?
(A) 1 (B) 2 (C) 3 (D) 4 (E) 5
A: 4.

Given mammogram image \textbf{<img>}. Image view: bilateral craniocaudal
Q: What is the most likely breast BI-RADS score?
(A) 1 (B) 2 (C) 3 (D) 4 (E) 5
A:
& 
3.\\
\midrule
\raisebox{-0.7\height}{\includegraphics[scale=0.3]{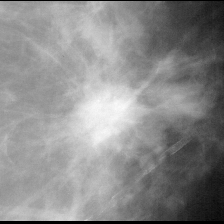}} 
& \textbf{Classification (CBIS-DDSM Calcification)}

Given mammogram image \textbf{<img>}. Image view: CC
Q: Which of the following is the most likely type of the patient's breast calcification?
(A) BENIGN (B) BENIGN\_WITHOUT\_CALLBACK (C) MALIGNANT

A:
& MALIGNANT. \\

\midrule
\raisebox{-0.6\height}{\includegraphics[scale=0.3]{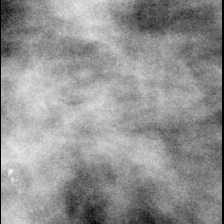}} 
& \textbf{Classification (CBIS-DDSM Mass)}

Given mammogram image \textbf{<img>}. Image view: CC
Q: Which of the following is the most likely type of the patient's breast mass? (A) BENIGN (B) BENIGN\_WITHOUT\_CALLBACK (C) MALIGNANT
A:
& BENIGN. \\
\midrule

\raisebox{-0.9\height}{\includegraphics[scale=0.3]{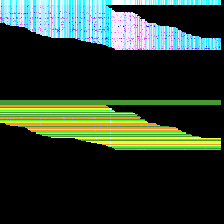}} 
& \textbf{Genomic variant calling} 

\textbf{Instructions:} You are a helpful genetic assistant. The following are questions about variant calling. Identify the number of copies of the putative variant in pileup images.

Given \textbf{<img>}. Q: How many copies of this putative variant are shown in the middle of the image? (A) 0 (B) 1 (C) 2 A:
& 
1.\\
\bottomrule
\end{tabular}
\label{tab:examples--multimedbench-classification}
\end{table}

\begin{table}[ht]
\footnotesize
\caption{\textbf{Examples of VQA and chest X-ray report generation tasks in MultiMedBench.}}
\centering
\begin{tabular}{p{0.15\linewidth} p{0.55\linewidth}  p{0.25\linewidth}}
\toprule
Image & Task and input prompt & Target\\ \midrule

\raisebox{-1\height}{\includegraphics[scale=0.3]{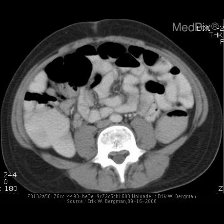}} 
& \textbf{VQA-RAD}

\textbf{Instructions:} You are a helpful medical assistant. The following are questions about medical knowledge. Solve them in a step-by-step fashion, referring to authoritative sources as needed.

Given <img>. Q: Can you diagnose a pericardial effusion from this image? (closed domain) 

A: No.

Given \textbf{<img>}. Q: What cut of the body is this image? (open domain) A:
& Axial. \\

\midrule
\raisebox{-1\height}{\includegraphics[scale=0.3]{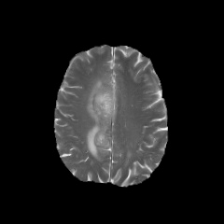}} 
& \textbf{Slake-VQA}

\textbf{Instructions:} You are a helpful medical assistant. The following are questions about medical knowledge. Solve them in a step-by-step fashion, referring to authoritative sources as needed.

Given <img>. Q: Is the lung healthy? 

A: No.

Given \textbf{<img>}. Q: Which part of the body does this image belong to? A:
& Brain. \\
\midrule

\raisebox{-1.2\height}{\includegraphics[scale=0.3]{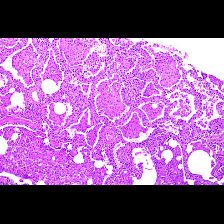}} 
& \textbf{Path-VQA} 

\textbf{Instructions:} You are a helpful medical assistant. The following are questions about medical knowledge. Solve them in a step-by-step fashion, referring to authoritative sources as needed.

Given <img>. Q: What is present ? (other) 

A: Abdomen.

Given \textbf{<img>}. Q: What is there of large numbers of macrophages within the alveolar spaces with only slight fibrous thickening of the alveolar walls? (other) 

A:
& 
accumulation of large numbers of macrophage.\\
\midrule

\raisebox{-2.3\height}{\includegraphics[scale=0.3]{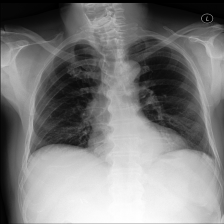}} 
& \textbf{Chest X-ray report generation} 

\textbf{Instructions:} You are a helpful radiology assistant. Describe what lines, tubes, and devices are present and each of their locations. Describe if pneumothorax is present; if present, describe size on each side. Describe if pleural effusion is present; if present, describe amount on each side. Describe if lung opacity (atelectasis, fibrosis, consolidation, infiltrate, lung mass, pneumonia, pulmonary edema) is present; if present, describe kinds and locations. Describe the cardiac silhouette size. Describe the width and contours of the mediastinum. Describe if hilar enlargement is present; if enlarged, describe side. Describe what fractures or other skeletal abnormalities are present. 

Given the LATERAL view X-ray image <img>. Reason for the study: Amiodarone routine surveillance. Q: Describe the findings in the image following the instructions.

A: As compared to the previous radiograph there is no relevant change. Normal lung volumes. Mild bilateral apical scarring. Normal size of the cardiac silhouette and tortuosity of the thoracic aorta. No pathologic findings in the lung parenchyma notably no evidence of fibrotic lung parenchymal changes. A faint 2 mm rounded opacity projecting over the lower aspect of the fourth right rib and internally to the upper border of the second right rib is seen on the frontal radiograph only and likely reflects structure on the skin. 

Given the PA view X-ray image \textbf{<img>}. Reason for the study: History m with malaise pneumonia. 
Q: Describe the findings in the image following the instructions. 

A:
& 
Again demonstrated is subtly increased opacity at the base of the right lung similar in appearance to multiple prior radiographs. There is no pneumothorax or pleural effusion. The cardiomediastinal and hilar contours are stable.\\
\bottomrule
\end{tabular}
\label{tab:examples--multimedbench-vqa-report-gen}
\end{table}

\begin{table}[!]
\footnotesize
\centering
\caption{\textbf{Example of the radiology report summarization task in MultiMedBench.}}
\vspace{3pt}
\label{tab:examples--report-summarization}
\begin{tabular}{l@{\hspace{.1em}}l@{\hspace{0.1em}}}
\toprule
{{\begin{tabular}[l]{@{}p{0.98\textwidth}}
\textbf{\textit{Input}}

\textbf{Instructions:} You are a helpful radiology assistant. The following are questions about radiology reports. Summarize the findings in the report into diagnostic statements. 

Given the findings: there is an intraparenchymal hemorrhage in the right cerebellar hemisphere measuring 1.7 cm with vasogenic edema and mass effect to the fourth ventricle. there is high density within the fissure of the right cerebellum suggestive of subarachnoid component. there is high density along the right tentorium, possibly representing subarachnoid hematoma, however, the finding is equivocal. there is no hydrocephalus, but there is mass effect and distortion of the fourth ventricle. there is no shift of normally midline supratentorial structures, and \underline{\hspace{0.6cm}}- white differentiations are preserved in the cerebral hemisphere. the surrounding osseous and soft tissue structures are unremarkable. mastoid air cells are not well pneumatized. there is mild mucosal thickening in the ethmoid sinuses. 
Q: Summarize the findings. 

A:
\\\\
\textbf{\textit{Target}}

1. 1.7-cm right cerebellar parenchymal hemorrhage with surrounding vasogenic edema and mass effect to the fourth ventricle, with adjacent subarachnoid hemorrhage. possible right subdural hemorrhage along the right tentorium, however, the evaluation is limited. differential diagnosis of the etiology of the bleeding included tumor, avm, and hypertension. the finding was discussed with dr. \underline{\hspace{0.6cm}} by telephone immediately after interpretation.
\end{tabular}}} 
& \\
\bottomrule 
 
\end{tabular}
\end{table}

\begin{table}[!]
\footnotesize
\centering
\caption{\textbf{Example from MedMCQA in MultiMedBench.}}
\vspace{3pt}
\label{tab:examples--medmcqa}
\begin{tabular}{l@{\hspace{.1em}}l@{\hspace{0.1em}}}
\toprule
{{\begin{tabular}[l]{@{}p{0.98\textwidth}}
\textbf{\textit{Input}}

\textbf{Instructions:} The following are multiple choice questions about medical knowledge. Solve them in a step-by-step fashion, starting by summarizing the available information. Output a single option from the four options as the final answer.
\\\\
Question: Which of the following is an intermediate-acting local anaesthetic which is an amino amide causing methemoglobinemia?\\
(A) Procaine (B) Prilocaine (C) Etidocaine (D) Ropivacaine\\
Answer: Prilocaine.
\\\\
Question: A 5-day-old male infant is diagnosed with Hirschsprung disease. CT scan examination reveals an abnormally dilated colon. Which of the following is the most likely embryologic mechanism responsible for Hirschsprung disease?\\
(A) Failure of neural crest cells to migrate into the walls of the colon (B) Incomplete separation of the cloaca (C) Failure of recanalization of the colon (D) Defective rotation of the hindgut\\
Answer: Failure of neural crest cells to migrate into the walls of the colon.
\\\\
Question: Chronic urethral obstruction due to benign prismatic hyperplasia can lead to the following change in kidney parenchyma
(A) Hyperplasia (B) Hyperophy (C) Atrophy (D) Dyplasia\\
Answer:
\\\\
\textbf{\textit{Target}}

Atrophy.
\end{tabular}}} 
& \\

\bottomrule 
\end{tabular}
\end{table}

\begin{table}[!]
\footnotesize
\centering
\caption{\textbf{Example from MedQA in MultiMedBench.}}
\vspace{3pt}
\label{tab:examples--medqa}
\begin{tabular}{l@{\hspace{.1em}}l@{\hspace{0.1em}}}
\toprule
{{\begin{tabular}[l]{@{}p{0.98\textwidth}}
\textbf{\textit{Input}}

\textbf{Instructions:} The following are multiple choice questions about medical knowledge. Solve them in a step-by-step fashion, starting by summarizing the available information. Output a single option from the four options as the final answer.\\
Question: A 57-year-old man presents to his family physician for a checkup. He has had type 2 diabetes mellitus for 13 years, for which he has been taking metformin and vildagliptin. He has smoked 10–15 cigarettes daily for 29 years. Family history is irrelevant. Vital signs include: temperature 36.6°C (97.8°F), blood pressure 152/87 mm Hg and pulse 88/min. Examination reveals moderate abdominal obesity with a body mass index of 32 kg/m². The remainder of the examination is unremarkable. His fasting lipid profile is shown: Total cholesterol (TC) 280 mg/dL Low-density lipoprotein (LDL)-cholesterol 210 mg/dL High-density lipoprotein (HDL)-cholesterol 40 mg/dL Triglycerides (TGs) 230 mg/dL Which of the following is the mechanism of action of the best initial therapy for this patient?\\
(A) Inhibition of cholesterol absorption (B) Bile acid sequestration (C) Inhibition of cholesterol synthesis (D) Activation of PPAR-alpha\\
Answer: Inhibition of cholesterol synthesis.\\ \\
Question: A 3-year-old girl presents with her mother for a well-child checkup. Recent laboratory data has demonstrated a persistent normocytic anemia. Her mother denies any previous history of blood clots in her past, but she says that her mother has also had to be treated for pulmonary embolism in the recent past, and her brother has had to deal with anemia his entire life. The patient's past medical history is noncontributory other than frequent middle ear infections. The vital signs upon arrival include: temperature, 36.7°C (98.0°F); blood pressure, 106/74 mm Hg; heart rate, 111/min and regular; and respiratory rate, 17/min. On physical examination, her pulses are bounding and fingernails are pale, but breath sounds remain clear. Oxygen saturation was initially 91\% on room air and electrocardiogram (ECG) shows sinus tachycardia. The patient's primary care physician orders a peripheral blood smear to further evaluate this finding, and preliminary results show a hemolytic anemia. Which of the following pathophysiologic mechanisms best describes sickle cell disease?\\
(A) Increased red blood cell sensitivity to complement activation, making patients prone to thrombotic events (B) A recessive beta-globin mutation causing morphological changes to the RBC (C) An X-linked recessive disease in which red blood cells are increasingly sensitive to oxidative stress (D) Secondarily caused by EBV, mycoplasma, CLL, or rheumatoid disease\\
Answer: A recessive beta-globin mutation causing morphological changes to the RBC.\\ \\
Question: A pulmonary autopsy specimen from a 58-year-old woman who died of acute hypoxic respiratory failure was examined. She had recently undergone surgery for a fractured femur 3 months ago. Initial hospital course was uncomplicated, and she was discharged to a rehab facility in good health. Shortly after discharge home from rehab, she developed sudden shortness of breath and had cardiac arrest. Resuscitation was unsuccessful. On histological examination of lung tissue, fibrous connective tissue around the lumen of the pulmonary artery is observed. Which of the following is the most likely pathogenesis for the present findings?\\
(A) Thromboembolism (B) Pulmonary ischemia (C) Pulmonary hypertension (D) Pulmonary passive congestion\\
Answer:
\\\\
\textbf{\textit{Target}}

Thromboembolism.
\end{tabular}}} 
& \\

\bottomrule 
 
\end{tabular}
\end{table}

\end{refsection}
\end{document}